\documentclass[final]{cvpr}

\usepackage{times}
\usepackage{epsfig}
\usepackage{graphicx}
\usepackage{amsmath}
\usepackage{amssymb}

\usepackage{comment}
\usepackage{color}
\usepackage[font=scriptsize]{subfig}
\usepackage{tabularx}
\usepackage{booktabs}
\usepackage{multirow}
\usepackage{multicol}
\usepackage{makecell}

\usepackage[pagebackref=true,breaklinks=true,colorlinks,bookmarks=false]{hyperref}

\newcommand{\mX}{\mathcal{X}}
\newcommand{\mU}{\mathcal{U}}
\newcommand{\norm}[1]{\left\lVert#1\right\rVert}

\def\cvprPaperID{85} 
\def\confYear{CVPR 2021}

\begin{document}

\title{Boosting Unconstrained Face Recognition with Auxiliary Unlabeled Data}

\author{Yichun Shi \quad\quad  Anil K. Jain\\[10pt]
Michigan State University\\[5pt]
{\tt\small shiyichu@msu.edu, jain@cse.msu.edu}
}

\maketitle

\begin{abstract}
In recent years, significant progress has been made in face recognition, which can be partially attributed to the availability of large-scale labeled face datasets. However, since the faces in these datasets usually contain limited degree and types of variation, the resulting trained models generalize poorly to more realistic unconstrained face datasets. While collecting labeled faces with larger variations could be helpful, it is practically infeasible due to privacy and labor cost. In comparison, it is easier to acquire a large number of unlabeled faces from different domains, which could be used to regularize the learning of face representations. We present an approach to use such unlabeled faces to learn generalizable face representations, where we assume neither the access to identity labels nor domain labels for unlabeled images. Experimental results on unconstrained datasets show that a small amount of unlabeled data with sufficient diversity can (i) lead to an appreciable gain in recognition performance and (ii) outperform the supervised baseline when combined with less than half of the labeled data. Compared with the state-of-the-art face recognition methods, our method further improves their performance on challenging benchmarks, such as IJB-B, IJB-C and IJB-S. 
\end{abstract}

\vspace{-1.0em}
\section{Introduction}
\label{sec:intro}
Machine learning algorithms typically assumes that training and testing data come from the same underlying distribution. However, in practice, we would often encounter testing domains that are different from the population where the training data is drawn. Since it is non-trivial to collect data for all possible testing domains, learning representations that are generalizable to heterogeneous testing data is desired~\cite{muandet2013domain,ghifary2015domain,motiian2017unified,li2018domain,carlucci2019domain}. Particularly for face recognition, this problem is reflected by the domain gap between the semi-constrained training datasets and unconstrained testing datasets. Nearly all of the state-of-the-art deep face networks are trained on large-scale web-crawled face images, most of which are high-quality celebrity photos~\cite{yi2014learning,guo2016msceleb}. But in practice, we wish to deploy the trained FR systems for many other scenarios, e.g. unconstrained photos~\cite{IJBA,IJBC} and surveillance~\cite{IJBS}. The large degree of face variation in the testing scenarios, compared to the training set, could result in significant performance drop of the trained face models~\cite{IJBC,IJBS}. 

\begin{figure}[t]
\captionsetup{font=footnotesize}
    \centering
    \includegraphics[width=1.00\linewidth]{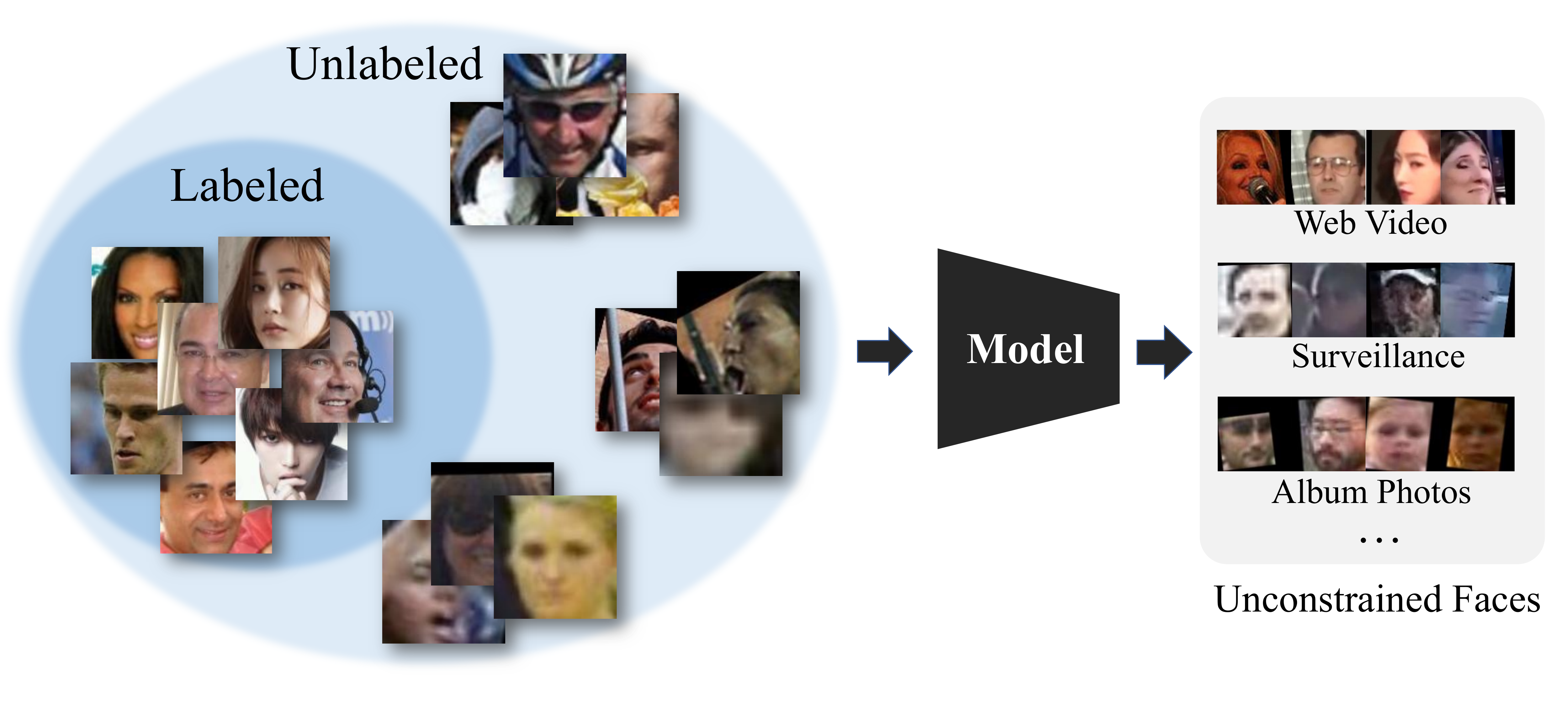}
    \vspace{-1.0em}\caption{Illustration of the problem settings in our work. Blue circles imply the domains that training face images belong to. By utilizing diverse unlabeled images, we want to regularize the learning of the face embedding for more unconstrained face recognition scenarios.}\vspace{-1.0em}
    \label{fig:problem_setting}
\end{figure}

The simplest solution to such a domain gap problem is to collect a large number of unconstrained labeled face images from different sources. However, due to privacy issue and human-labeling cost, it is extremely hard to collect such a database. Other popular solutions to this problem include transfer learning and domain adaptation, which require domain-specific data to train a model for each of the target domains~\cite{pan2010domain,ganin2014unsupervised,long2017deep,sohn2017unsupervised,saito2018maximum,kang2019contrastive}. However, in unconstrained face recognition, a face representation that is robust to all different kinds of variations is needed, so these domain-specific solutions are not appropriate. \emph{Instead, it would be useful if we could utilize the commonly available, unlabeled data to achieve a domain-agnostic face representation that generalizes to unconstrained testing scenarios} (See Fig.~\ref{fig:problem_setting}). To achieve this goal, we would like to ask the following questions in this paper:
\begin{itemize}\vspace{-0.5em}
    \item Is it possible to improve model generalizability to unconstrained faces by introducing more diversity from auxiliary unlabeled data?\vspace{-0.5em}
    \item What kind of and how much unlabeled data do we need?\vspace{-0.5em}
    \item How much performance boost could we achieve with the unlabeled data?\vspace{-0.5em}
\end{itemize}


In this paper, we propose such an semi-supervised framework for learning robust face representations. The unlabeled images are collected from a public face detection  dataset, i.e. WiderFace~\cite{yang2016wider}, which contains more diverse types (sub-domains) of face images compared to typical labeled face datasets used for training. 
To utilize the unlabeled data, the proposed method jointly regularizes the embedding model from feature space and image space. We show that adversarial regularization can help to reduce domain gaps caused by facial variations, even in the absence of sub-domain labels. On the other hand, an image augmentation module is trained to discover the hidden sub-domain styles in the unlabeled data and apply them to the labeled training samples, thus increasing the discrimination power on difficult face examples. To our knowledge, this is the first study to use a heterogeneous unlabeled dataset to boost the model performance for general unconstrained face recognition. The contributions of this paper are summarized as below:
\begin{itemize}\vspace{-0.5em}
    \item A semi-supervised learning framework for generalizing face representations with auxiliary unlabeled data.\vspace{-0.5em}
    \item An multi-mode image translation module is proposed to perform data-driven augmentation and increase the diversity of the labeled training samples.\vspace{-0.5em}
    \item Empirical results show that the regularization of unlabeled data helps to improve the recognition performance on challenging testing datasets, e.g. IJB-B, IJB-C, and IJB-S.
\end{itemize}

\section{Related Work}
\subsection{Deep Face Recognition}
Deep neural networks are widely adopted in the ongoing research in face recognition ~\cite{taigman2014deepface,deepid2,schroff2015facenet,masi2016we,liu2017sphereface,hasnat2017deepvisage,ranjan2017l2,wang2018additive,deng2018arcface}. Taigman et al.~\cite{taigman2014deepface} were the first to propose using deep convolutional neural network for learning face representations. The subsequent studies have explored different loss functions to improve the discrimination power of the learned feature representation. A number of studies proposed to use metric learning methods for face recognition ~\cite{schroff2015facenet,sohn2016improved}. Recent work has been trying to achieve discriminative embeddings with a single identification loss function where proxy/prototype vectors are used to represent each class in the embedding space~\cite{liu2017sphereface,wang2018additive,wang2018cosface,ranjan2017l2,deng2018arcface,zhang2019adacos,sun2020circle}.

\subsection{Semi-supervised Learning}
Classic semi-supervised learning involves a small number of labeled images and a large number of unlabeled images~\cite{lee2013pseudo,rasmus2015semi,laine2017temporal,tarvainen2017mean,xie2019unsupervised,zhai2019s4l,berthelot2019mixmatch,sohn2020fixmatch}. The goal is to improve the recognition performance when we don't have sufficient data that are labeled. State-of-the-art semi-supervised learning methods can mainly be classified into four categories. (1) Pseudo-labeling methods generate labels for unlabeled data with the trained model and then use them for training~\cite{lee2013pseudo}. In spite of its simplicity, it has been shown to be effective primarily for classification tasks where labeled data and unlabeled data share the same label space. (2) Temporal ensemble models maintain different versions of model parameters to serve as teacher models for the current model~\cite{laine2017temporal,tarvainen2017mean}. (3) Consistency-regularization methods apply certain types of augmentation to the unlabeled data while making sure the output prediction remains consistent after augmentation~\cite{rasmus2015semi,berthelot2019mixmatch,sohn2020fixmatch}. (4) Self-supervised learning, originally proposed for unsupervised learning, has recently been shown to be effective for semi-supervised learning as well~\cite{zhai2019s4l}. Compared with classic semi-supervised learning addressed in the literature, our problem is different in two sense of heterogeneity: different domains and different identities between the labeled and unlabeled data. These differences make many classic semi-supervised learning methods unsuitable for our task.

\subsection{Domain Adaptation and Generalization}
In domain adaptation, the user has a dataset for a source domain and another for a fixed target domain~\cite{pan2010domain,ganin2014unsupervised,long2017deep,saito2018maximum,kang2019contrastive}. If the target domain is unlabeled, this leads to an \emph{unsupervised domain adaption} setting~\cite{ganin2014unsupervised,tzeng2017adversarial,saito2018maximum,kang2019contrastive}. The goal is to improve the performance on the target domain so that it could match the performance on the source domain. This is achieved by reducing the domain gap between the two datasets in feature space. The problem about domain adaption is that one needs to acquire a new dataset and train a new model whenever there is a new target domain. In \emph{domain generalization}, the user is given a set of labeled datasets from different domains. The model is jointly trained on these datasets so that it could better generalize to unseen domains~\cite{muandet2013domain,ghifary2015domain,motiian2017unified,li2018domain,carlucci2019domain,guo2020learning}. Our problem lies in the middle between domain generalization and unsupervised domain adaptation : we want to generalize the model to broader domains, yet instead of multi-domain labeled data, we use unlabeled data from other sources to achieve this goal.

\begin{figure}[t]
    \centering
    \captionsetup{font=footnotesize}
    \scriptsize
    \includegraphics[width=1.00\linewidth]{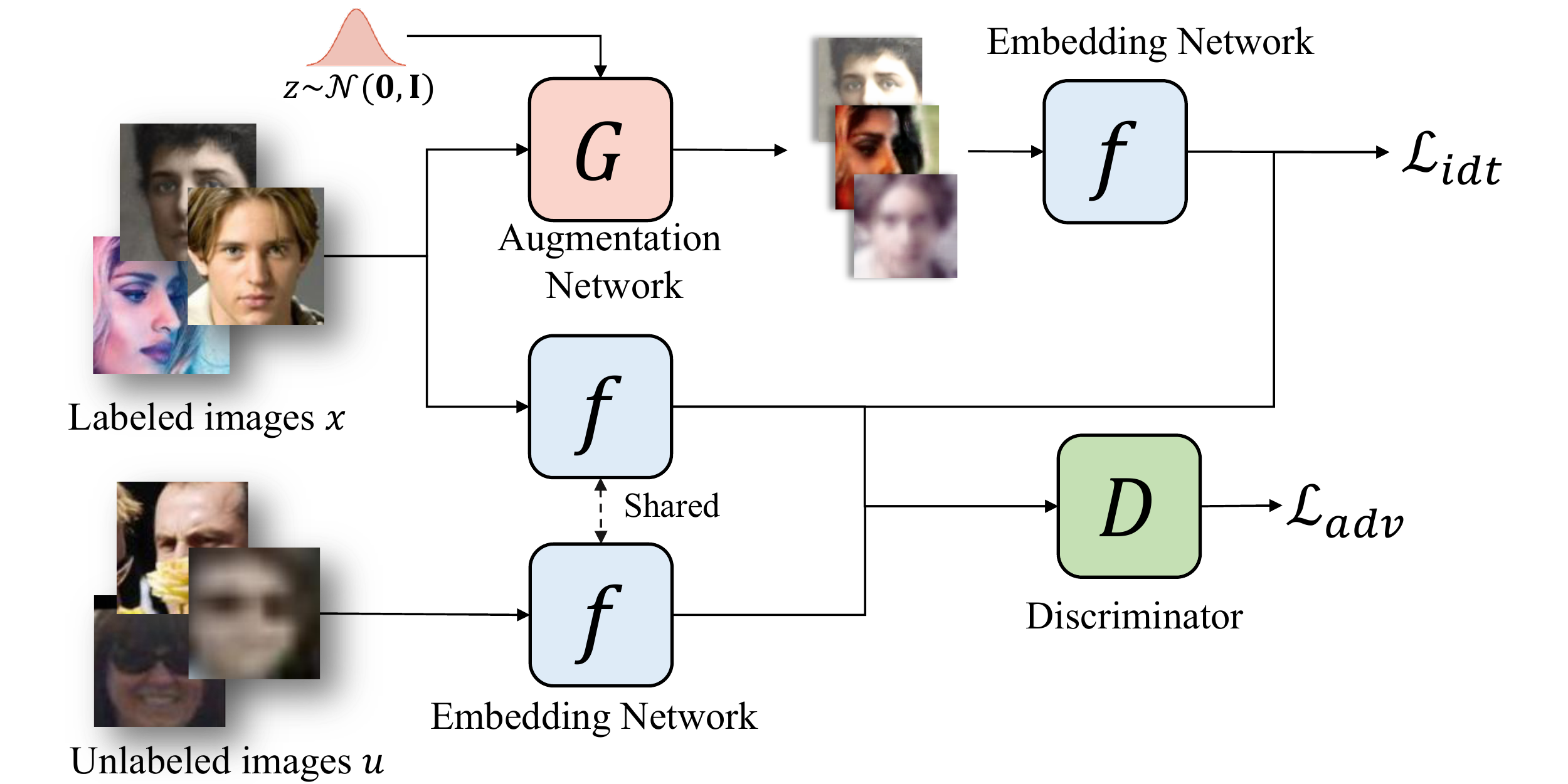}
    \caption{The training framework of the embedding network. In each mini-batch, a random subset of labeled data would be augmented by the augmentation network to introduce additional diversity. The non-augmented labeled data are used to train the feature discriminator. The adversarial loss forces the distribution of the unlabeled features to align with the labeled one.}
    \label{fig:overview}
\end{figure}

\section{Methodology}
\label{sec:method}
Generally, in face representation learning, we are given a large labeled dataset $\mX$=$\{(x_1,y_1),(x_2,y_2),\dots,(x_n,y_n)\}$, where $x_i$ and $y_i$ are the face images and identity labels, respectively. The goal is to learn an embedding model $f$ such that $f(x)$ would be discriminative enough to distinguish between different identities. However, since $f$ is only trained on the domain defined by $\mX$, which is usually semi-constrained celebrity photo, it might not generalize to unconstrained settings. In our framework, we assume the availability of another unlabeled dataset $\mU=\mU_1\cup \mU_2\dots\mU_k=\{u_1,u_2,\dots,u_n\}$,
collected from different sources (sub-domains). However, these sub-domain labels may not be available in real applications, thus we do not assume the access to them but instead seek solutions that could automatically leverage these hidden sub-domains. 
Then, we wish to simultaneously minimize three types of errors:
\begin{itemize}
    \item Error due to discrimination power within the labeled domain $\mX$.
    \item Error due to feature domain gap between the labeled domain $\mX$ and the hidden sub-domains $\mU_i$.
    \item Error due to discrimination power within the unlabeled domain $\mU$.
\end{itemize}
An overview of the framework is shown in Fig.~\ref{fig:overview}.

\subsection{Minimizing Error in the Labeled Domain}
\label{sec:method_labeled}
The deep representation of a face image is usually a point in a hyper-spherical embedding space, where $\norm{f(x_i)}^2=1$. State-of-the-art supervised face recognition methods all try to find an objective function to maximize the inter-class margin such that the representation could still be discriminative when tested on unseen identities. In this work, we choose to use CosFace loss function~\cite{wang2018cosface}\cite{wang2018additive} for training the labeled images:
\begin{equation}
    \mathcal{L}_{idt} = -\mathbb{E}_{x_i,y_i\sim\mX}[\log \frac{e^{s(W_{y_i}^Tf_i-m)}}{e^{s(W_{y_i}^Tf_i-m)}+\sum_{j\neq y_i}{e^{sW_{y_j}^Tf_i}}}].
\end{equation}
Here $s$ is the hyper-parameter controlling temperature, $m$ is a margin hyper-parameter and $W_{j}$ is the proxy vector of the $j^{th}$ identity in the embedding space, which is also $\ell_2$ normalized. We choose to use CosFace loss function because of its stability and high-performance. It could potentially be replaced by any other supervised identification loss function.

\begin{figure}[t]
    \captionsetup{font=footnotesize}
    \centering
    \subfloat[w/o Domain Adversarial Loss]{\includegraphics[width=0.48\linewidth]{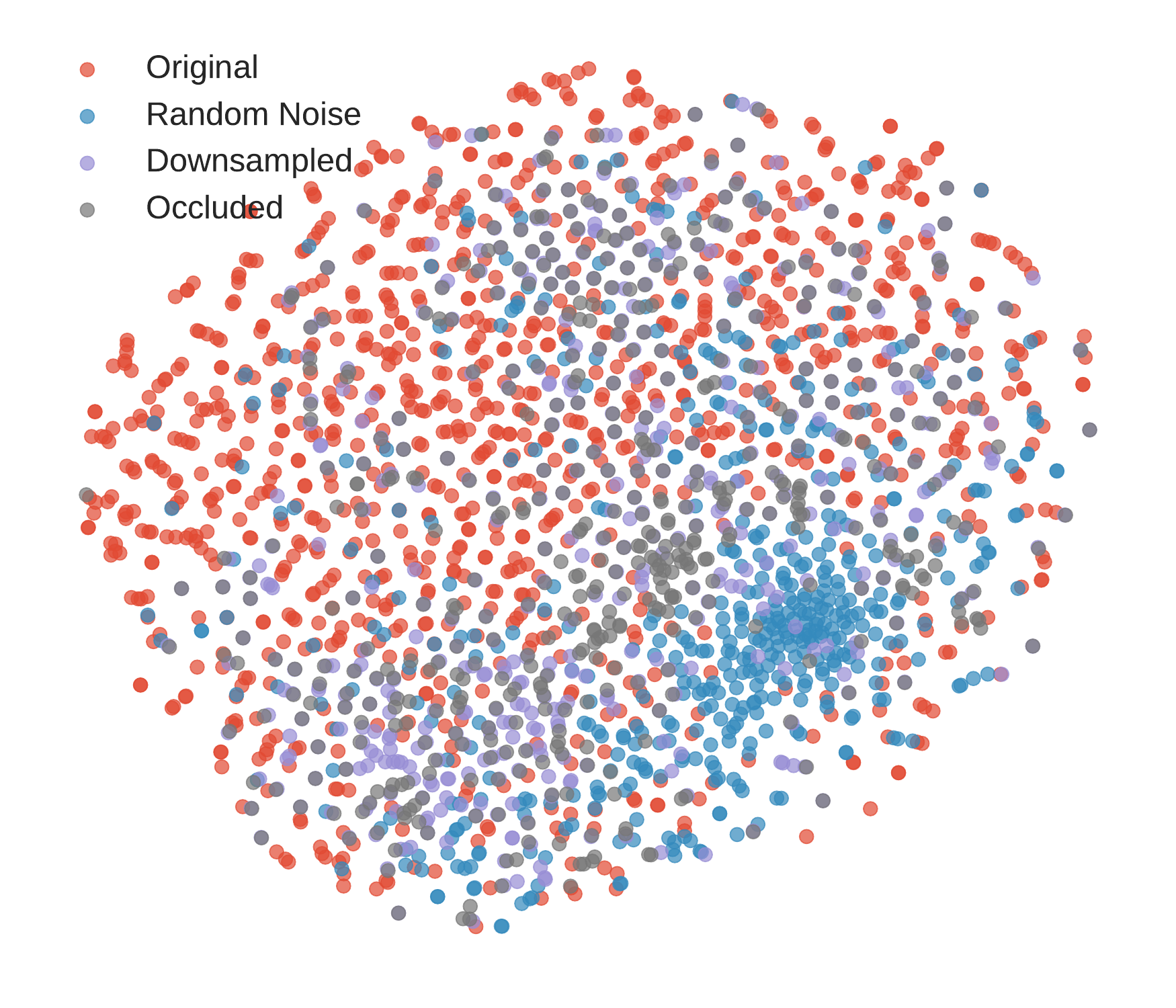}}\hfill
    \subfloat[w/ Domain Adversarial Loss]{\includegraphics[width=0.48\linewidth]{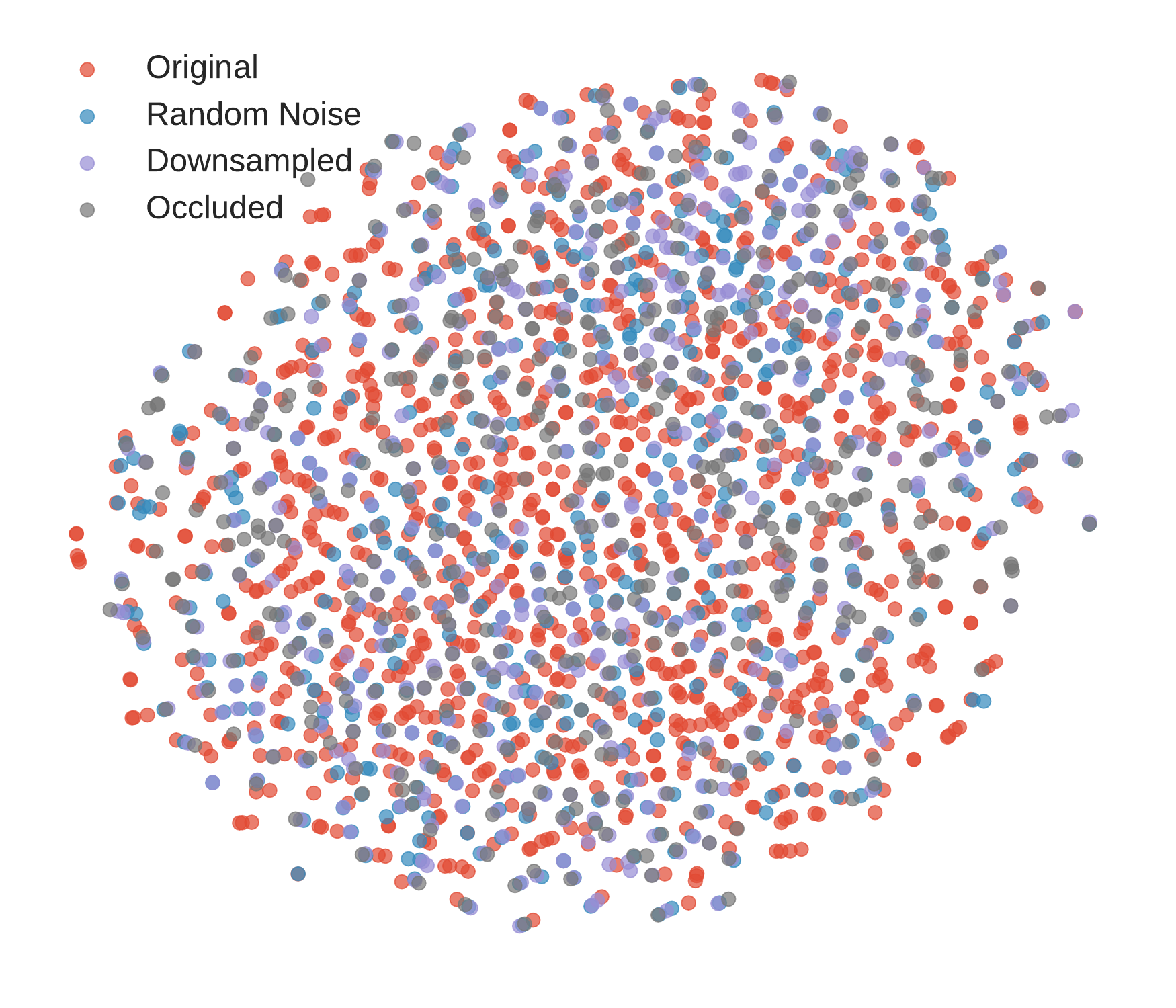}}
    \caption{t-SNE visualization of the face embeddings using synthesized unlabeled images. Using part of the MS-Celeb-1M as unlabeled dataset, we create three sub domains by processing the images with either random Gaussian noise, random occlusion or downsampling. (a) different sub-domains show different domain shift in the embedding space of the supervised baseline. (b) with the holistic binary domain adversarial loss, each of the sub-domains is aligned with the distribution of the labeled data.}
    \label{fig:tsne_subdomain}
\end{figure}

\subsection{Minimizing Domain Gaps}
\label{sec:method_domaingap}
The unlabeled dataset $\mU$ is assumed to be a diverse dataset collected from different sources, i.e. covering different sub-domains (types) of face images. If we have the access to such sub-domain labels, a natural solution to a domain-agnostic model would be aligning each of the sub-domains with the feature distribution of the labeled images. However, the sub-domain labels might not be available in many cases. In our experiment, we find there is no necessity for pairwise domain alignment. Instead, a binary domain alignment loss is sufficient to align the sub-domains. Formally, given a feature discriminator network $D$, we could reduce the domain gap via an adversarial loss:
\begin{gather}
\begin{split}
    \mathcal{L}_{D} = -\mathbb{E}_{x\sim\mX}[\log D(y=0|f(x)]\\
    -\mathbb{E}_{u\sim\mU}[\log D(y=1|f(u)],
\end{split}\\
\begin{split}
    \mathcal{L}_{adv} = -\mathbb{E}_{x\sim\mX}[\log D(y=1|f(x)]\\
    -\mathbb{E}_{u\sim\mU}[\log D(y=0|f(u)].
\end{split}
\label{eq:adv}
\end{gather}
The discriminator $D$ is a multi-layer binary classifier optimized by $\mathcal{L}_{D}$. It tries to learn a non-linear classification boundary between the two datasets while the embedding network needs to fool the discriminator by reducing the divergence between the distributions of $f(x)$ and $f(u)$. 
To see the effect of domain alignment loss, we conduct a controlled experiments with a toy dataset. We split the MS-Celeb-1M~\cite{guo2016msceleb} dataset into labeled images and unlabeled images (no identity overlap). The unlabeled images are then processed with one of the three degradations: random Gaussian noise, random occlusion and downsampling. Thus, we create three sub-domains in the unlabeled dataset. The corresponding domain shift can be observed in the t-SNE plot in Fig.~\ref{fig:tsne_subdomain} (a), where the model is trained only on the labeled split. Then, we incorporate the augmented unlabeled images into training with the binary domain adversarial loss. In Fig.~\ref{fig:tsne_subdomain} (b), we observe that with the binary domain alignment loss, the distribution of each of sub-domains is aligned with the original domain, indicating reduced domain gaps.

\subsection{Minimizing Error in the Unlabeled Domains}
\label{sec:method_unlabeled}


The domain alignment loss in Section~\ref{sec:method_domaingap} helps to eliminate the error caused by domain gaps between unconstrained faces. Thus, the remaining task is to improve the discrimination power of the face representation among the unlabeled faces. Many semi-supervised classification methods address this problem by using pseudo-labeling of unlabeled data~\cite{lee2013pseudo,berthelot2019mixmatch,sohn2020fixmatch}, but this is not applicable to our problem since our unlabeled dataset does not share the same label space with the labeled one. Furthermore, because of data collection protocols, there is very little chance that one identity would have multiple unlabeled images. Thus, clustering-based methods are also infeasible for our task. Here, we consider to address this issue with a multi-mode augmentation method.
Prior studies have shown that an image translation network, such as CycleGAN~\cite{zhu2017unpaired}, can be effectively used as a data augmentation module for domain adaptation~\cite{hoffman2018cycada}. The main idea of the augmentation network is to learn the difference between two domains in the image space and then augment the samples from source domain data to create training data with pseudo-labels in the target domain. Since our goal is to generalize the deep face representation to unconstrained faces, which involves a large variety, deterministic method such as CycleGAN would be unsuitable. Therefore, we propose to use a multi-mode image translation network that could discover the hidden domains in the unlabeled data and then augment the labeled training data with different styles. In particular, we need a function $G$ which maps labeled samples $x$ into the image space defined by the unlabeled faces, i.e. $p(x)\rightarrow p(u)$. Then, training the embedding $f$ on $G(x)$ could make it more discriminative in the image space defined by $U$. There are two requirement of the function $G$: (1) it should not change the identity of the input image and (2) it should be able to capture different styles that are present in the unlabeled images. Inspired by recent progress in image translation frameworks~\cite{zhu2017unpaired,MUNIT}, we propose to train $G$ as a style-transfer network that learns the visual styles during transfer in an unsupervised manner. The network $G$ can then be used as a data-driven augmentation module that generates diverse samples given an input from the labeled dataset. During the training, we randomly replace a subset of the labeled images to be augmented and put them into our identification learning framework. The details of training the augmentation network $G$ is given in Section~\ref{sec:generator}.

The overall loss function for the embedding network is given by:
\begin{equation}
    \mathcal{L} = \lambda_{idt}\mathcal{L}_{idt} + \lambda_{adv}\mathcal{L}_{adv}
\end{equation}
where $L_{idt}$ also includes the augmented labeled samples.

\subsubsection{Multi-mode Augmentation Network}
\label{sec:generator}

\begin{figure}[t]
\captionsetup{font=footnotesize}
    \centering
    \scriptsize
    \includegraphics[width=1.00\linewidth]{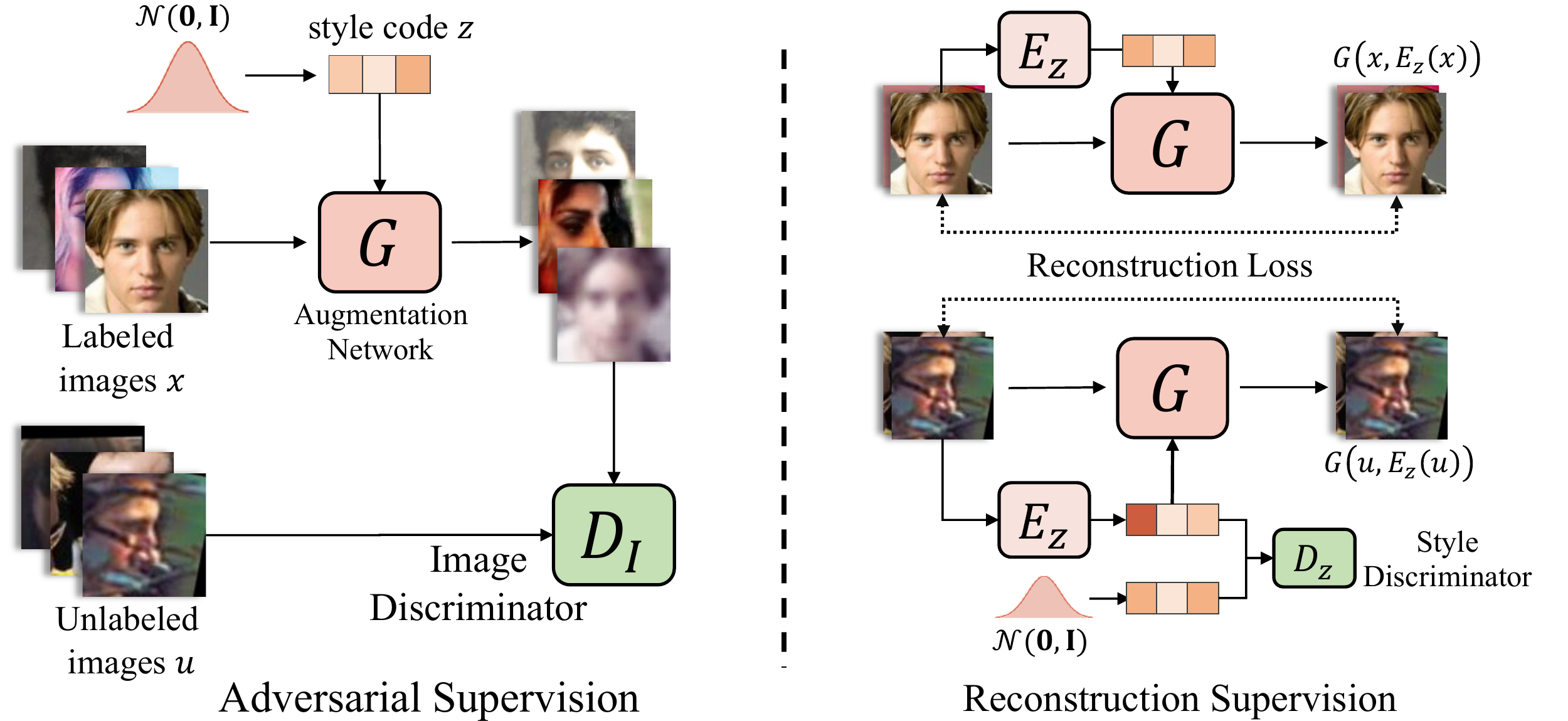}
    \caption{Training framework of the augmentation network $G$. The two pipelines are optimized jointly during training.}
    \label{fig:overview_generator}
\end{figure}

The augmentation network $G$ is a fully convolutional network that maps an image to another. To preserve the geometric structure, our architecture does not involve any downsampling or upsampling. In order to generate styles similar to the unlabeled images, an image discriminator $D_I$ is trained to distinguish between the texture styles of unlabeled images and generated images:
\begin{align}
\begin{split}
    \mathcal{L}_{D_I} = & -\mathbb{E}_{x\sim\mX}[\log D_I(y=0|G(x,z))]\\
    & -\mathbb{E}_{u\sim\mU}[\log D_I(y=1|u)], 
\end{split}\\
    \mathcal{L}^{G}_{adv} = & -\mathbb{E}_{x\sim\mX}[\log D_I(y=1|G(x,z))].
\end{align}
Here $z\sim\mathcal{N}(\mathbf{0},\mathbf{I})$ is a random style vector to control the styles of the output image, which is injected into the generation process via Adaptive Instance Normalization (AdaIN)~\cite{huang2017adain}.  Although the adversarial learning could make sure the output are in the unlabeled space, but it cannot ensure that (1) the content of the input is maintained in the output image and (2) the random style $z$ is being used to generate diverse visual styles, corresponding to differnt sub-domains in the unlabeled images. We propose to utilize an additional reconstruction pipeline to simultaneously satisfy these two requirements. First, we introduce an additional style encoder $E_z$ to capture the corresponding style in the input image, as in~\cite{MUNIT}. A reconstruction loss is then enforced to keep the consistency of the image content:
\begin{align}
    \mathcal{L}^{G}_{rec} = & \; \mathbb{E}_{x\sim\mX}[\norm{x-G(x,E_z(x))}^2] \\
    & + \mathbb{E}_{u\sim\mU}[\norm{u-G(u,E_z(u))}^2],
\end{align}
Then, during the reconstruction, we add another latent style discriminator $D_z$ to guarantee the distribution of $E_z(u)$ align with prior distribution $\mathcal{N}(\mathbf{0},\mathbf{I})$:
\begin{align}
\begin{split}
    \mathcal{L}_{D_z} = & -\mathbb{E}_{u\sim\mU}[\log D_z(y=0|E_z(u))]\\
    & -\mathbb{E}_{z\sim\mathcal{N}(\mathbf{0},\mathbf{I})}[\log D_z(y=1|z)], 
\end{split}\\
    \mathcal{L}^{z}_{adv} = & -\mathbb{E}_{u\sim\mU}[\log D_z(y=1|E_z(u))],
\end{align}

The overall loss function of the generator is given by:
\begin{equation}
    \mathcal{L}^{G} = \lambda^{G}_{adv}\mathcal{L}^{G}_{adv} + \lambda^{G}_{rec}\mathcal{L}^{G}_{rec} + \lambda^{z}_{adv}\mathcal{L}^{z}_{adv}
\end{equation}
A overview of the training framework of $G$ is given in Fig.~\ref{fig:overview_generator} and example generated images are shown in Fig.~\ref{fig:augmentation}. The architecture details of different modules are given in the supplementary file.

\begin{figure}[t]
    \centering
    \footnotesize
    \captionsetup{font=footnotesize}
    \begin{minipage}{1.0\linewidth}
    \includegraphics[width=0.14\linewidth]{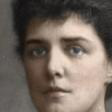}\hfill
    \includegraphics[width=0.14\linewidth]{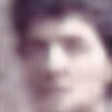}\hfill
    \includegraphics[width=0.14\linewidth]{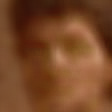}\hfill
    \includegraphics[width=0.14\linewidth]{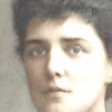}\hfill
    \includegraphics[width=0.14\linewidth]{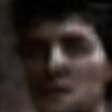}\hfill
    \includegraphics[width=0.14\linewidth]{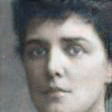}\hfill
    \includegraphics[width=0.14\linewidth]{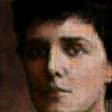}\hfill\\[-0.05em]
    \includegraphics[width=0.14\linewidth]{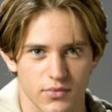}\hfill
    \includegraphics[width=0.14\linewidth]{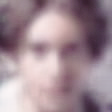}\hfill
    \includegraphics[width=0.14\linewidth]{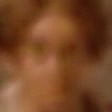}\hfill
    \includegraphics[width=0.14\linewidth]{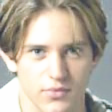}\hfill
    \includegraphics[width=0.14\linewidth]{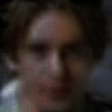}\hfill
    \includegraphics[width=0.14\linewidth]{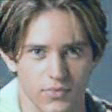}\hfill
    \includegraphics[width=0.14\linewidth]{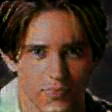}\hfill\\[-0.05em]
    \includegraphics[width=0.14\linewidth]{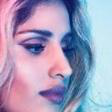}\hfill
    \includegraphics[width=0.14\linewidth]{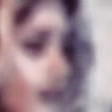}\hfill
    \includegraphics[width=0.14\linewidth]{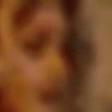}\hfill
    \includegraphics[width=0.14\linewidth]{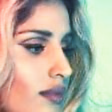}\hfill
    \includegraphics[width=0.14\linewidth]{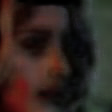}\hfill
    \includegraphics[width=0.14\linewidth]{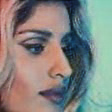}\hfill
    \includegraphics[width=0.14\linewidth]{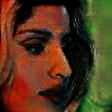}\hfill\\[-0.05em]
    \end{minipage}\hfill
    \vspace{-1.0em}\caption{Example generated images of the augmentation network. Each row shows augmented images with different styles for the input in the first column.}
    \label{fig:augmentation}\vspace{-1.0em}
\end{figure}

\section{Experiments}

\subsection{Implementation Details}

\noindent\textbf{Training Details of the Recognition Models}
All the models are implemented with Pytorch v1.1. We use the RetinaFace~\cite{deng2019retinaface} for face detection and alignment. All images are transformed into $112\times112$ pixels. A modified 50-layer ResNet in~\cite{deng2018arcface} is used as our architecture. The embedding size is $512$ for all models. By default, all the models are trained with $150,000$ steps with a batch size of 256. For semi-supervised models, we use $64$ unlabeled images and $192$ labeled images in each mini-batch. For models which use the augmentation module, $20\%$ of the labeled images are augmented by the generator network. The scale parameter $s$ and margin parameter $m$ are set to $30$ and $0.5$, respectively. We empirically set $\lambda_{idt}$, $\lambda_{adv}$ as 1.0 and 0.01. 

\noindent\textbf{Training Details of the Generator Models}
The generator is trained for $160,000$ steps with a batch size of $8$ images ($4$ from each dataset). Adam optimizer is used with $\beta_1=0.5$ and $\beta_2=0.99$. The learning rate starts with $1e-4$ and drops to $1e-5$ after $80,000$ steps. The detailed architectures are provided in the supplementary material. $\lambda^{G}_{adv}$, $\lambda^{G}_{rec}$ and 
$\lambda^{z}_{adv}$ are set to as 1.0, 10.0 and 1.0, respectively.

\begin{figure}[t]
    \centering
    \footnotesize
    \captionsetup{font=footnotesize}
    \begin{minipage}{0.49\linewidth}
    \includegraphics[width=0.33\linewidth]{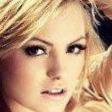}\hfill
    \includegraphics[width=0.33\linewidth]{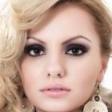}\hfill
    \includegraphics[width=0.33\linewidth]{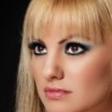}\hfill\\[-0.1em]
    \includegraphics[width=0.33\linewidth]{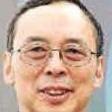}\hfill
    \includegraphics[width=0.33\linewidth]{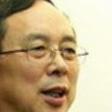}\hfill
    \includegraphics[width=0.33\linewidth]{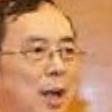}\hfill\\[-0.1em]
    \includegraphics[width=0.33\linewidth]{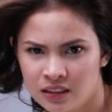}\hfill
    \includegraphics[width=0.33\linewidth]{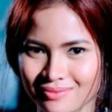}\hfill
    \includegraphics[width=0.33\linewidth]{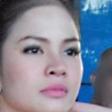}\hfill\\
    \vspace{-2.0em}\begin{center}(b) MsCeleb1M\end{center}
    \end{minipage}\hfill
    \begin{minipage}{0.49\linewidth}
    \includegraphics[width=0.33\linewidth]{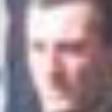}\hfill
    \includegraphics[width=0.33\linewidth]{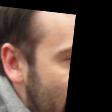}\hfill
    \includegraphics[width=0.33\linewidth]{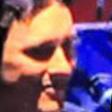}\hfill\\[-0.1em]
    \includegraphics[width=0.33\linewidth]{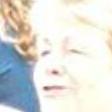}\hfill
    \includegraphics[width=0.33\linewidth]{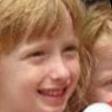}\hfill
    \includegraphics[width=0.33\linewidth]{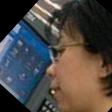}\hfill\\[-0.1em]
    \includegraphics[width=0.33\linewidth]{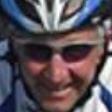}\hfill
    \includegraphics[width=0.33\linewidth]{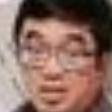}\hfill
    \includegraphics[width=0.33\linewidth]{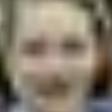}\hfill\\
    \vspace{-2.0em}\begin{center}(a) WiderFace\end{center}
    \end{minipage}\\
    \begin{minipage}{0.49\linewidth}
    \includegraphics[width=0.33\linewidth]{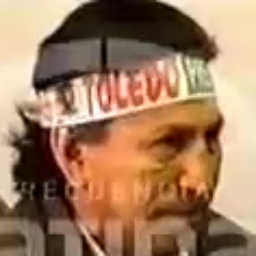}\hfill
    \includegraphics[width=0.33\linewidth]{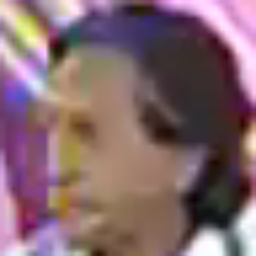}\hfill
    \includegraphics[width=0.33\linewidth]{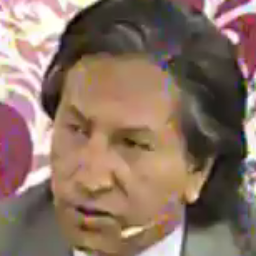}\hfill\\[-0.1em]
    \includegraphics[width=0.33\linewidth]{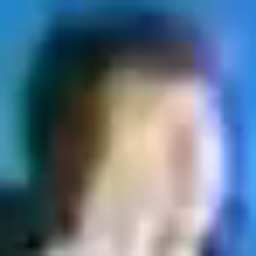}\hfill
    \includegraphics[width=0.33\linewidth]{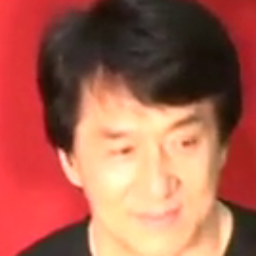}\hfill
    \includegraphics[width=0.33\linewidth]{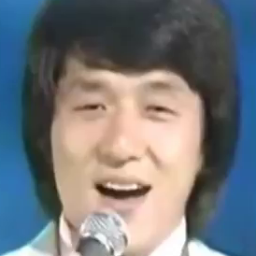}\hfill\\[-0.1em]
    \includegraphics[width=0.33\linewidth]{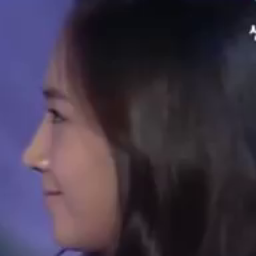}\hfill
    \includegraphics[width=0.33\linewidth]{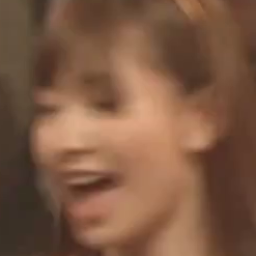}\hfill
    \includegraphics[width=0.33\linewidth]{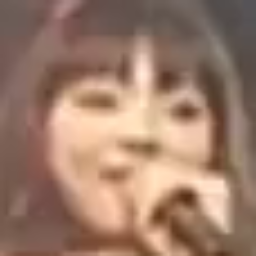}\hfill\\
    \vspace{-2.0em}\begin{center}(c) IJB-C\end{center}
    \end{minipage}\hfill
    \begin{minipage}{0.49\linewidth}
    \includegraphics[width=0.33\linewidth]{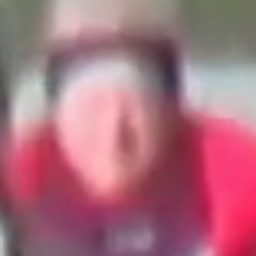}\hfill
    \includegraphics[width=0.33\linewidth]{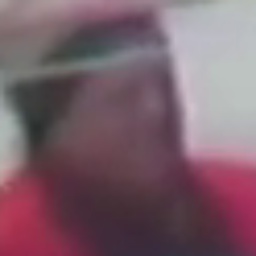}\hfill
    \includegraphics[width=0.33\linewidth]{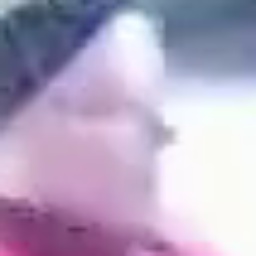}\hfill\\[-0.1em]
    \includegraphics[width=0.33\linewidth]{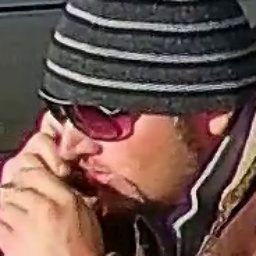}\hfill
    \includegraphics[width=0.33\linewidth]{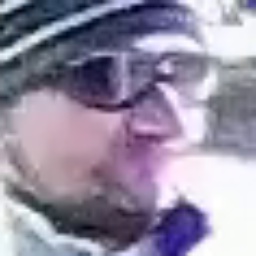}\hfill
    \includegraphics[width=0.33\linewidth]{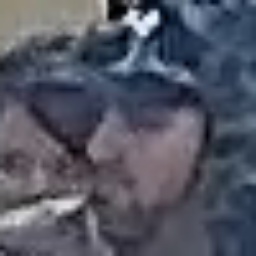}\hfill\\[-0.1em]
    \includegraphics[width=0.33\linewidth]{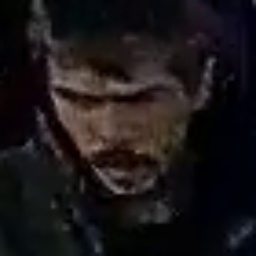}\hfill
    \includegraphics[width=0.33\linewidth]{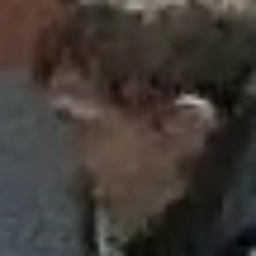}\hfill
    \includegraphics[width=0.33\linewidth]{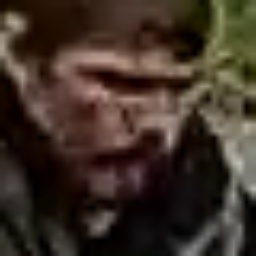}\hfill\\
    \vspace{-2.0em}\begin{center}(d) IJB-S\end{center}
    \end{minipage}\\
    \caption{Examples face images of different datasets used in this work. The unconstrained testing datasets (IJB-C, IJB-S) are significantly different from the labeled training data (MsCeleb) with more types and degree of variation. Thus, auxiliary data from WiderFace are utilized to generalize the model.}
    \label{fig:exp_dataset}\vspace{-1.0em}
\end{figure}

\subsection{Datasets}
We use \textbf{MS-Celeb-1M}~\cite{guo2016msceleb} as our labeled training dataset. MS-Celeb-1M is a large-scale public face dataset of celebrity photos. 
The original dataset is known to contain a large number of noisy labels~\cite{cao2018vggface2}, so we use a cleaned version from ArcFace~\cite{deng2018arcface} as training data. After removing the overlapped subjects with the testing sets and duplicate images, we are left with 3.9M images of 85.7K classes. As for unlabeled images, we choose \textbf{WiderFace}~\cite{yang2016wider} as our auxiliary training data. WiderFace is a dataset collected by retrieving images from search engines with different event keywords. As a face detection dataset, WiderFace includes a more diverse set of faces (See Fig.~\ref{fig:exp_dataset}). Many faces in this dataset still cannot be detected by state-of-the-art detection methods~\cite{deng2019retinaface}. Thus, we only keep the detectable faces in the WiderFace training set as our training data. Our goal is to close the gap between face detection and recognition engine and improve the general recognition performance for any detectable faces. At the end, we were able to detect about 70K faces from WiderFace, less than 2\% of our labeled training data. 

To evaluate the proposed method, we test on three unconstrained face recognition benchmarks, namely IJB-B, IJB-C and IJB-S.
These datasets represent real-world testing scenarios where faces are significantly different from celebrity photos in the training set. The details of these datasets are as follows:
\begin{itemize}\vspace{-0.5em}
    \item \textbf{IJB-B}~\cite{IJBB} includes both high quality celebrity photos taken from the wild and low quality photos or video frames with large variations of illumination, occlusion, head pose, etc. There are 68,195 images of 1,845 identities in all. We test on both verification and identification protocols of the IJB-B benchmark.\vspace{-0.5em}
    \item \textbf{IJB-C}~\cite{IJBC} is a newer version of IJB-B dataset. It has a similar protocol but with 140,732 images of 3,531 identities.\vspace{-0.5em}
    \item \textbf{IJB-S}~\cite{IJBS} is an extremely challenging benchmark where the images were collected from surveillance cameras. There are in all 202 identities with an average of 12 videos per person. Each person also has 7 high-quality enrollment photos (with different poses) which constitute the gallery. We test on two protocols of the IJB-S dataset, Surveillance-to-Still (\textbf{V2S}) and Surveillance-to-Booking (\textbf{V2B}), both of which are identification protocols. The difference between them is that in Surveillance-to-Still (V2S) the gallery of each person is a single frontal photo while Surveillance-to-Booking (V2B) uses all 7 registration photos as gallery. To Reduce the evaluation time, the expriments in Sec.~\ref{sec:exp_ablation} and Sec.~\ref{sec:exp_quantity} are conducted with subsampled frames from each video, whose performance is close to using the whole video (Sec.~\ref{sec:exp_final}).
\end{itemize}

Although our goal is to improve the recognition performance on unconstrained faces, we would not like to lose the discrimination power in the original domain (high-quality photos). Therefore, during ablation we also evaluate our models on the standard \textbf{LFW}~\cite{LFWTech} protocol, which is a celebrity photo dataset, similar to the labeled training data (MS-Celeb-1M). Note that the accuracy on the LFW protocol is highly saturated, so the main goal is just to check whether there is a significant performance drop on the constrained faces while increasing the generalizability to unconstrained ones. Example images of different datasets are shown in Figure~\ref{fig:exp_dataset}.

\subsection{Ablation Study}
\label{sec:exp_ablation}
In this section, we conduct an ablation study to quantitatively evaluate the effect of different modules proposed in this paper. In particular, we have two modules to study: Domain Alignment (DA) and Augmentation Network (AN). The performance is shown in Table~\ref{tab:ablation}. As we already showed in Fig.~\ref{fig:tsne_subdomain}, domain adversarial loss is able to force smaller domain gaps between the sub-domains in WiderFace and the celebrity faces, even though we do not have access to those domain labels. Consequently, we observe the performance improvement on most of the protocols on IJB-C and IJB-S. Introducing the augmentation network (AN) further helps improving the performance on unconstrained benchmarks, where a multi-mode (MM) augmentation network outperforms a single-model (SM) augmentation network. More details of ablating over the augmentation network can be found in the supplementatry material.

\begin{table}[t]
\captionsetup{font=footnotesize}
\setlength{\tabcolsep}{1.5pt}
\begin{center}
\scriptsize
\begin{tabularx}{1.00\linewidth}{X || c|c|c|c|c|c|c|c}
\Xhline{2\arrayrulewidth}
\multirow{2}{*}{Method}      & \multicolumn{3}{c|}{IJB-C (Vrf)} & \multicolumn{2}{c|}{IJB-C (Idt)} & \multicolumn{2}{c|}{IJB-S (V2S)} & LFW \\\cline{2-9}
                            & 1e-7 & 1e-6 & 1e-5 & Rank1 & Rank5& Rank1 & Rank5 & Accuracy \\
\Xhline{2\arrayrulewidth}
Baseline                    & 62.90 & 82.94 & 90.73 & 94.90 & 96.77 & 53.23 & 62.91 & 99.80 \\\hline
 + DA                       & 72.74 & 85.33 & 90.52 & 94.99 & 96.75 & 56.35 & \textbf{66.77} & \textbf{99.82} \\\hline
 + DA + AN (SM)             & 74.80 & 87.58 & \textbf{91.94} & 95.51 & 97.09 & 56.98 & 65.66 & 99.80 \\\hline
\textbf{ + DA + AN (MM)}    & \textbf{77.39} & \textbf{87.92} & 91.86 & \textbf{95.61} & \textbf{97.13} & \textbf{57.33} & 65.37 & 99.75 \\\hline
\Xhline{2\arrayrulewidth}
\end{tabularx}\vspace{0.5em}
\caption{Ablation study over different training methods of the embedding network. All models has identification loss by default. ``DA'', ``AN'', ``SM'' and ``MM'' refer to ``Domain Alignment'', ``Augmentation Network'', ``Single-mode'' and ``Multi-mode'', respectively.}\vspace{-1.0em}
\label{tab:ablation}
\end{center}
\end{table}

\subsection{Quantity vs. Diversity}
\label{sec:exp_quantity}

\begin{figure*}[t]
    \centering
    \includegraphics[width=0.33\linewidth]{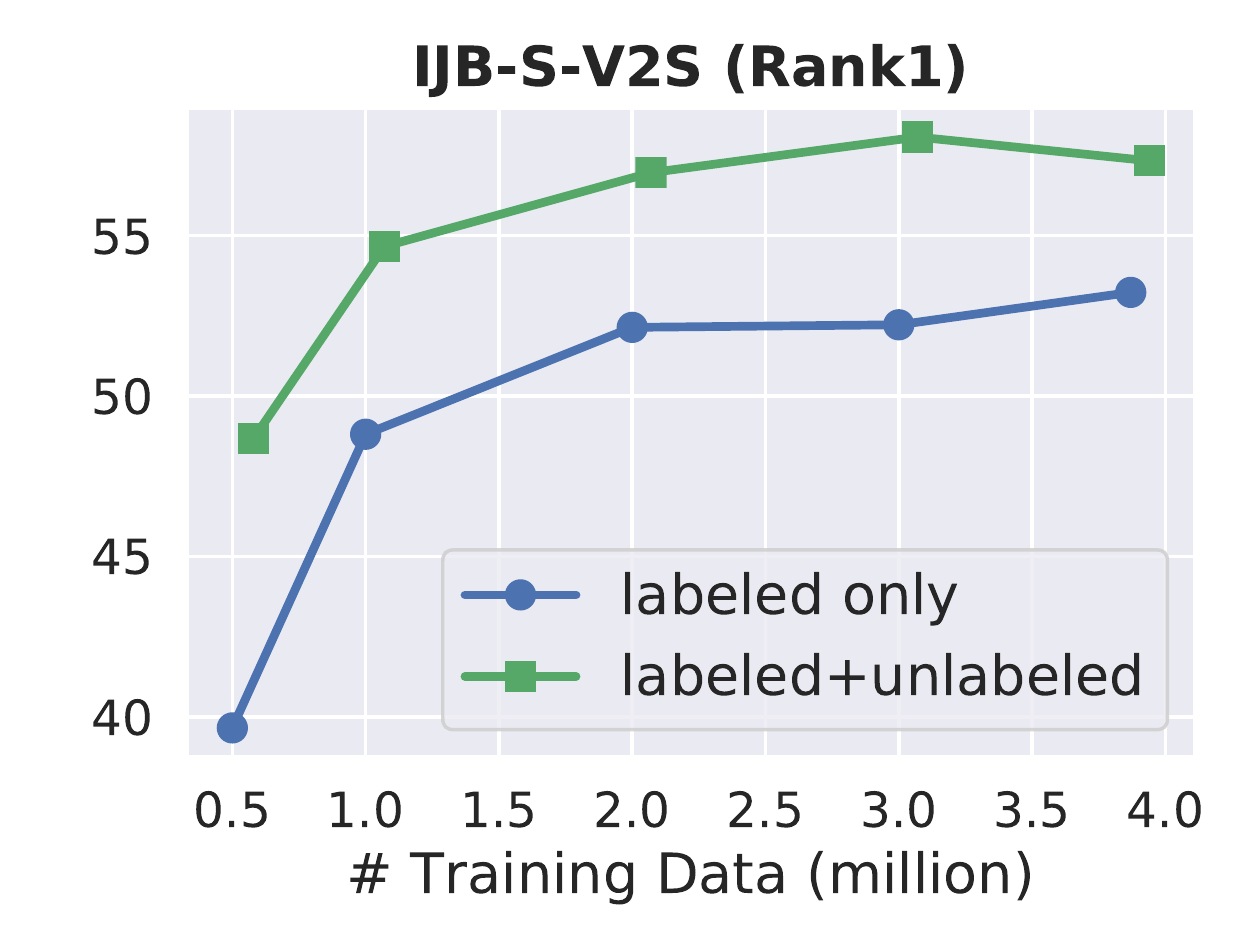}\hfill
    \includegraphics[width=0.33\linewidth]{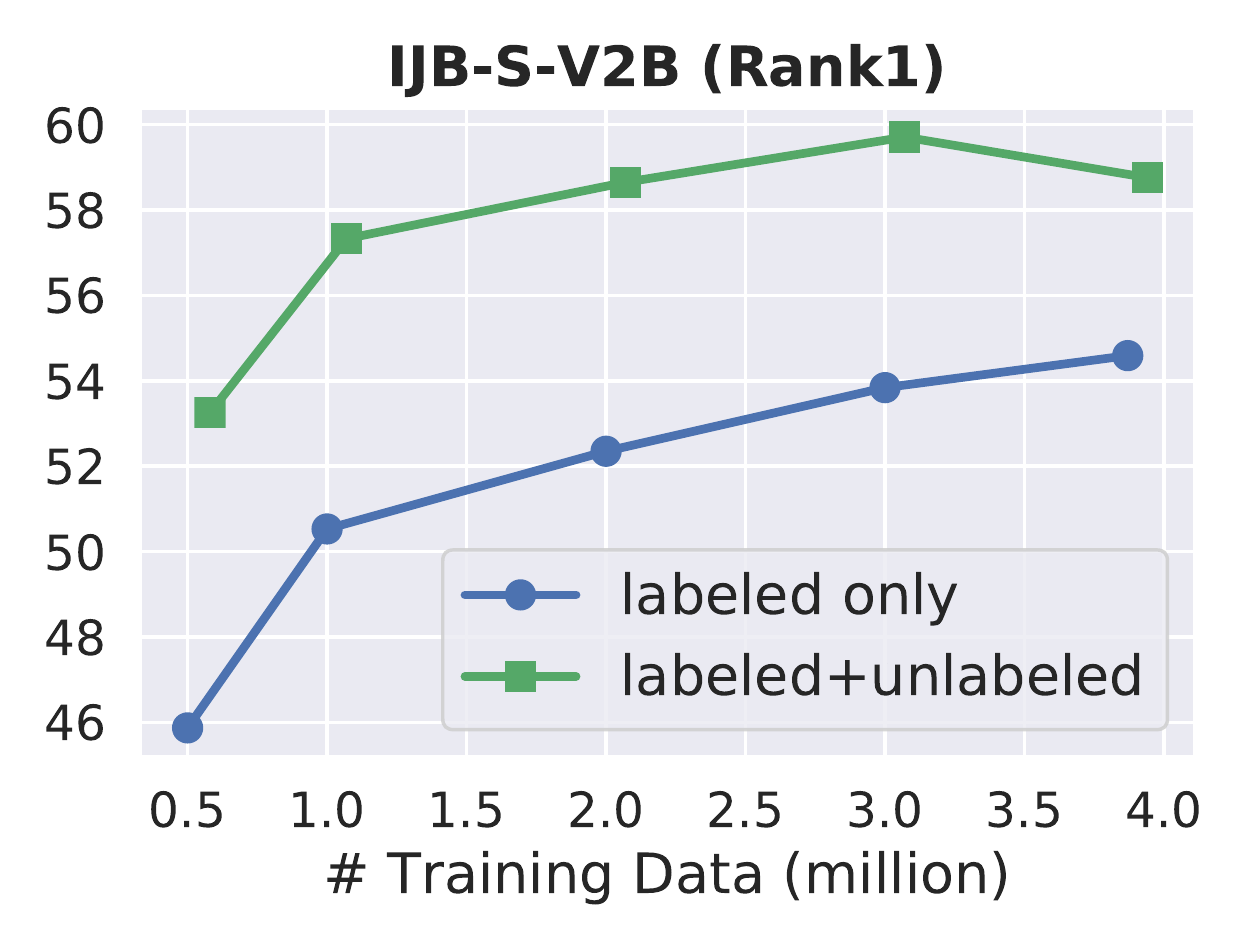}\hfill
    \includegraphics[width=0.33\linewidth]{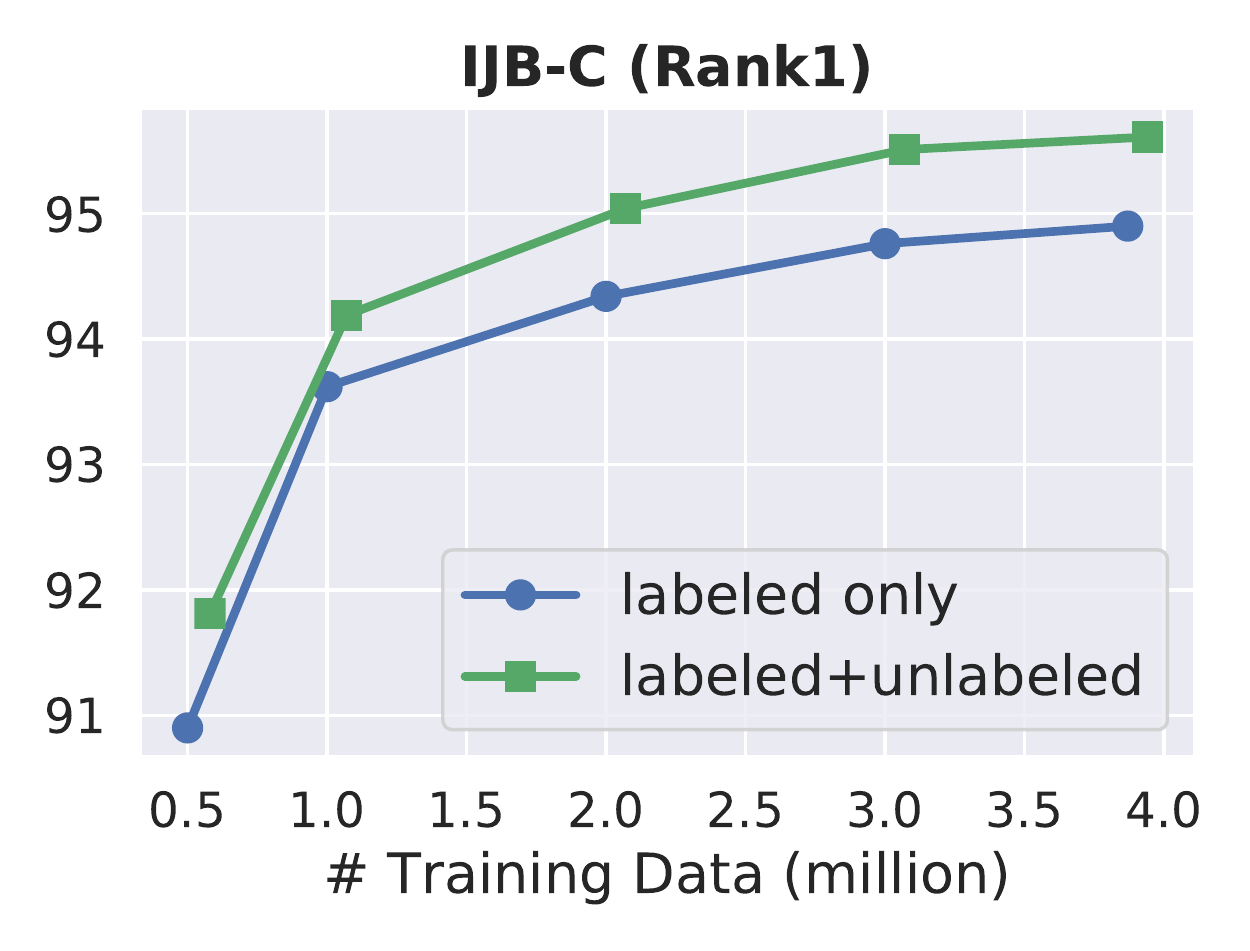}\\
    \includegraphics[width=0.33\linewidth]{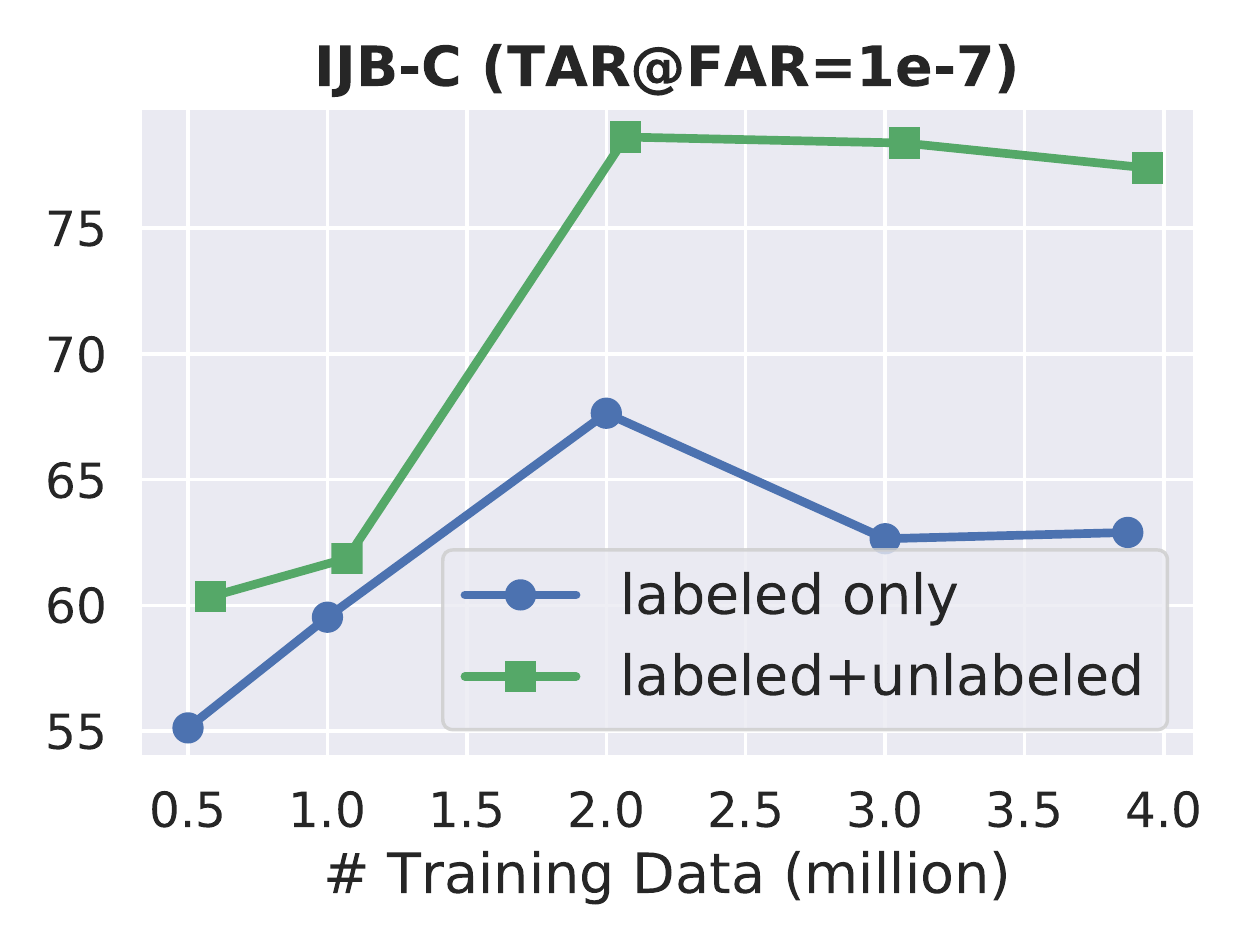}\hfill
    \includegraphics[width=0.33\linewidth]{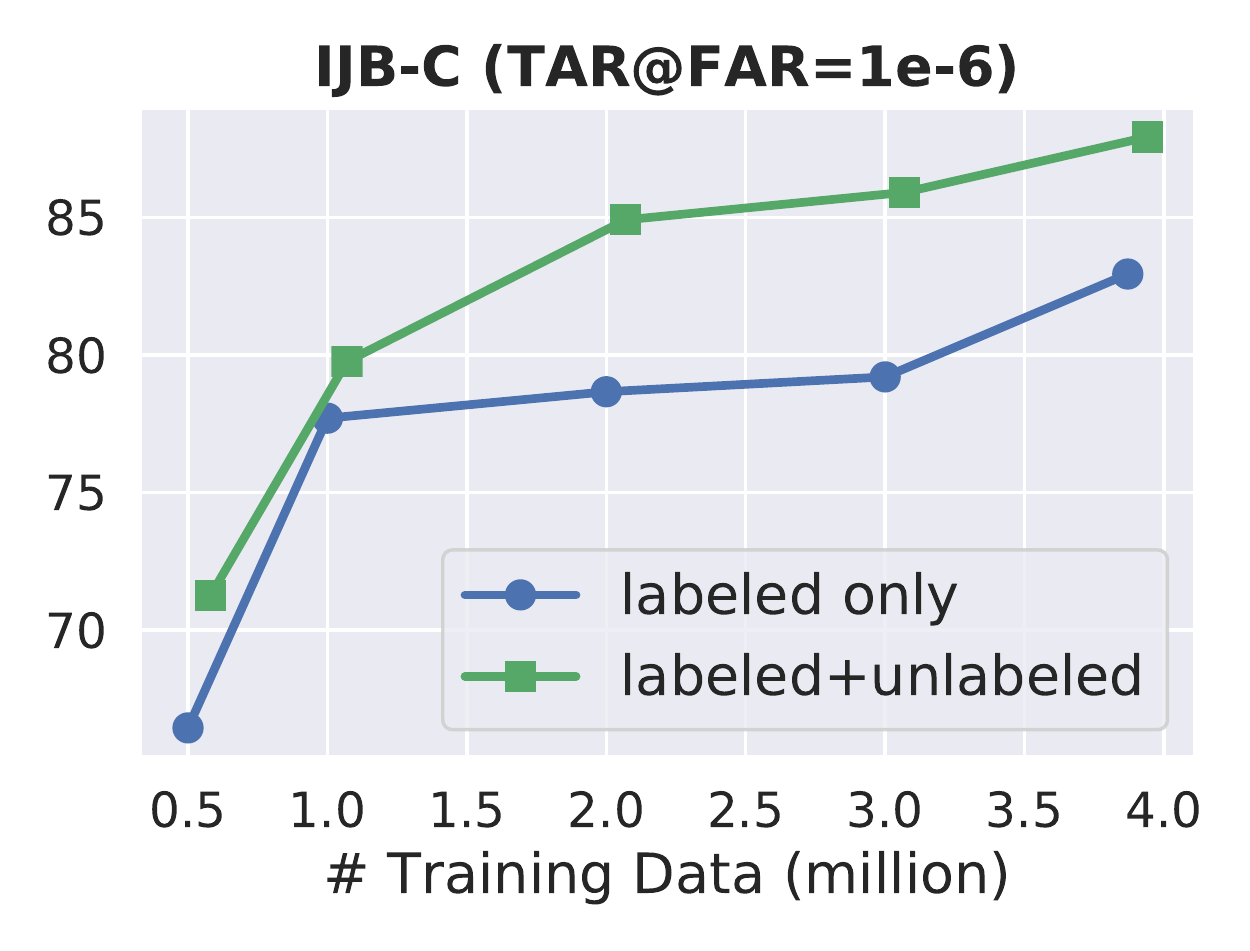}\hfill
    \includegraphics[width=0.33\linewidth]{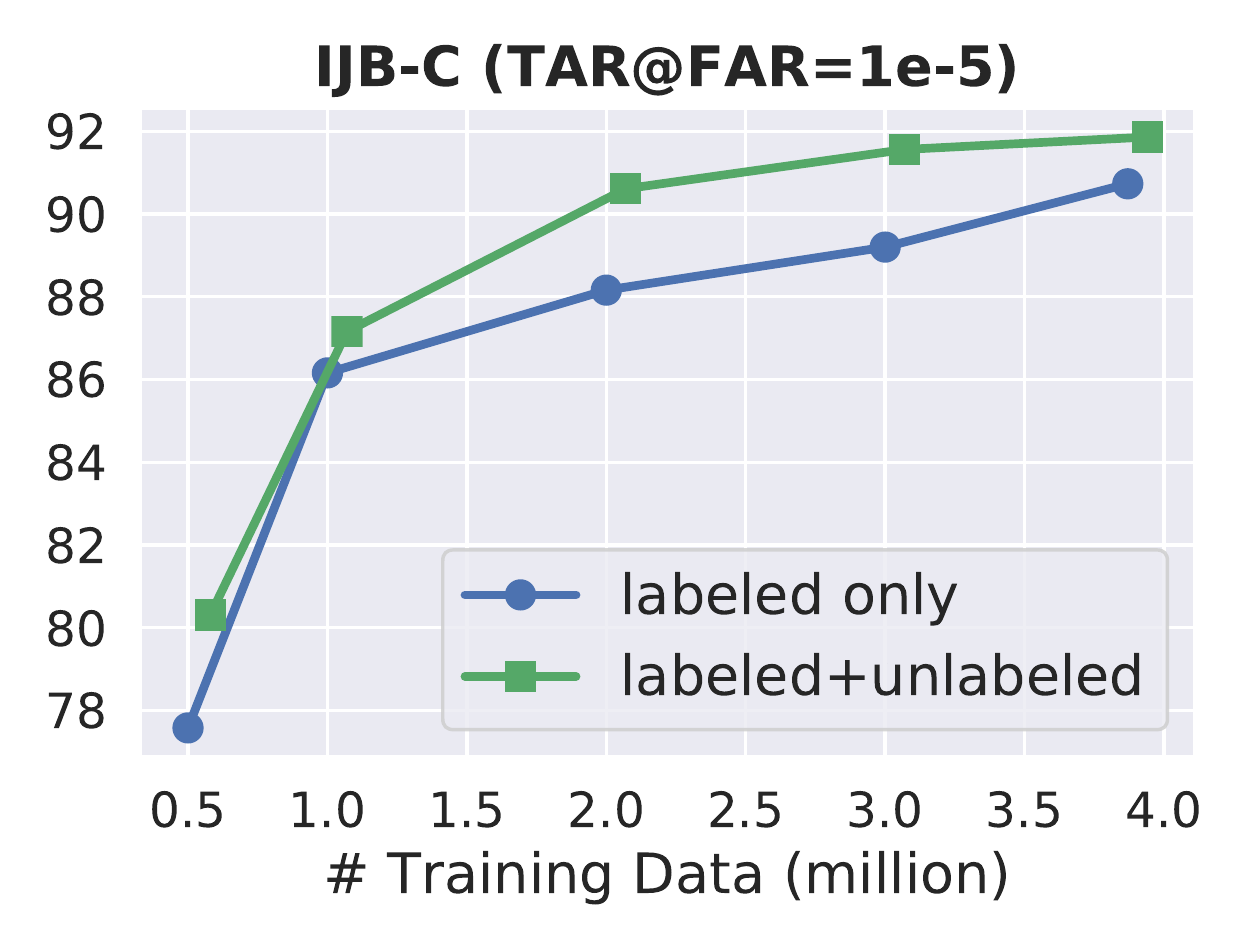}\\[-0.8em]
    \caption{Evaluation Results on IJB-C and IJB-S with different protocols and different number of labeled training data.}
    \label{fig:quanitity}
\end{figure*}

Although we have shown in Sec.~\ref{sec:exp_ablation} that utilizing unlabeled data leads to better performance on challenging testing benchmarks, generally it shall be expected that simply increasing the number of labeled training data can also have a similar effect. Therefore, in this section, we conduct a more detailed study to answer such a question: \textit{which is more important for feature generalizability: quantity or diversity of the training data?} In particular, we train several supervised models by adjusting the number of labeled training data. For each such model, we also train a corresponding model with additional unlabeled data. The evaluation results are shown in Figure~\ref{fig:quanitity}.

On the IJB-S dataset, which is significantly different from the labeled training data, we see that the models trained with unlabeled data consistently outperform the supervised baselines with a large margin. In particular, the proposed method achieves better performance than the supervised baseline even when there is only one-fourth of the overall labeled training data (1M vs 4M), indicating the value of data diversity during training. Note that there is a significant performance boost when increasing the number of labeled samples from 0.5M to 1M. However, after that, the benefit of acquiring more labeled data plateaus and in fact it is more helpful to introduce 70K unlabeled data than 3M additional labeled data.

On the IJB-C dataset, for both verification and identification protocols, we observe a similar trend as the IJB-S dataset. In particular, a larger improvement is achieved at lower FARs. This is because the verification threshold at lower FARs is affected by the low quality test data (difficult impostor pairs), which is more similar to our unlabeled data. Another interesting observation is that the improvement margin increases when there is more labeled data. Note that in general semi-supervised learning, we would expect less improvement by using unlabeled data when there is more labeled data. But it is the opposite in our case because the unlabeled data has different characteristics than the labeled data. So when the performance of supervised model saturates with sufficient labeled data, transferring the knowledge from diverse unlabeled data becomes more helpful.

For both IJB-S and IJB-C (TAR@FAR=1e-7), we observe that after a certain point, adding more labeled data does not boost performance any more and the performance starts to fluctuate. This happens because the new labeled data does not necessarily help with those hard cases. Based on these results, we conclude that \emph{when the number of labeled training data is small, it is more important to increase the quantity of the labeled dataset. Once there is sufficient labeled training data, the generalizablity of the representation tends to saturate while the diversity of the training data becomes more important}. 
Additional experimental results on the choice of the unlabeled dataset can be found in Section~\ref{sec:appendix_unlabeled}.

\begin{figure}[h]
\captionsetup{font=small}
    \centering
    \includegraphics[width=0.49\linewidth]{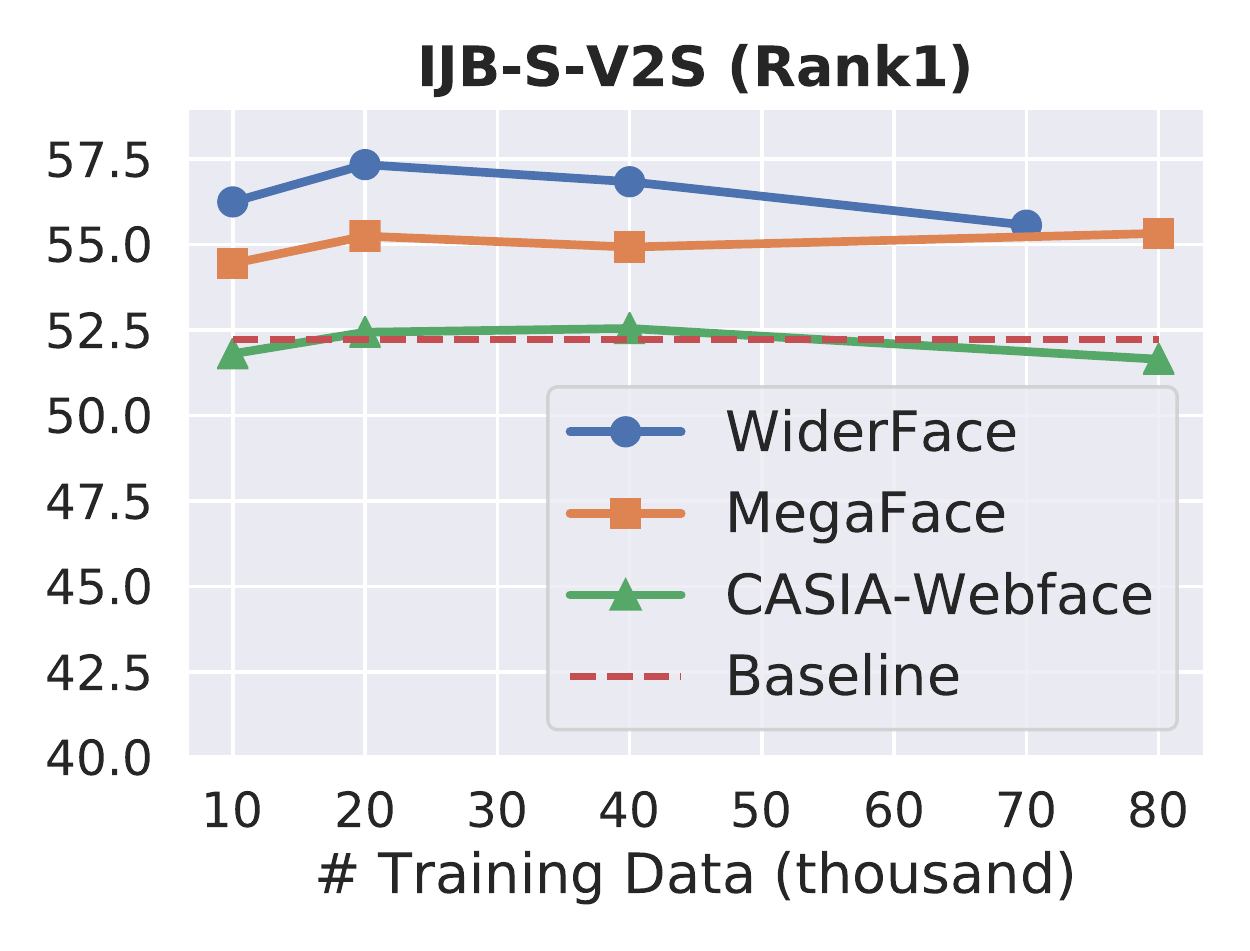}\hfill
    \includegraphics[width=0.49\linewidth]{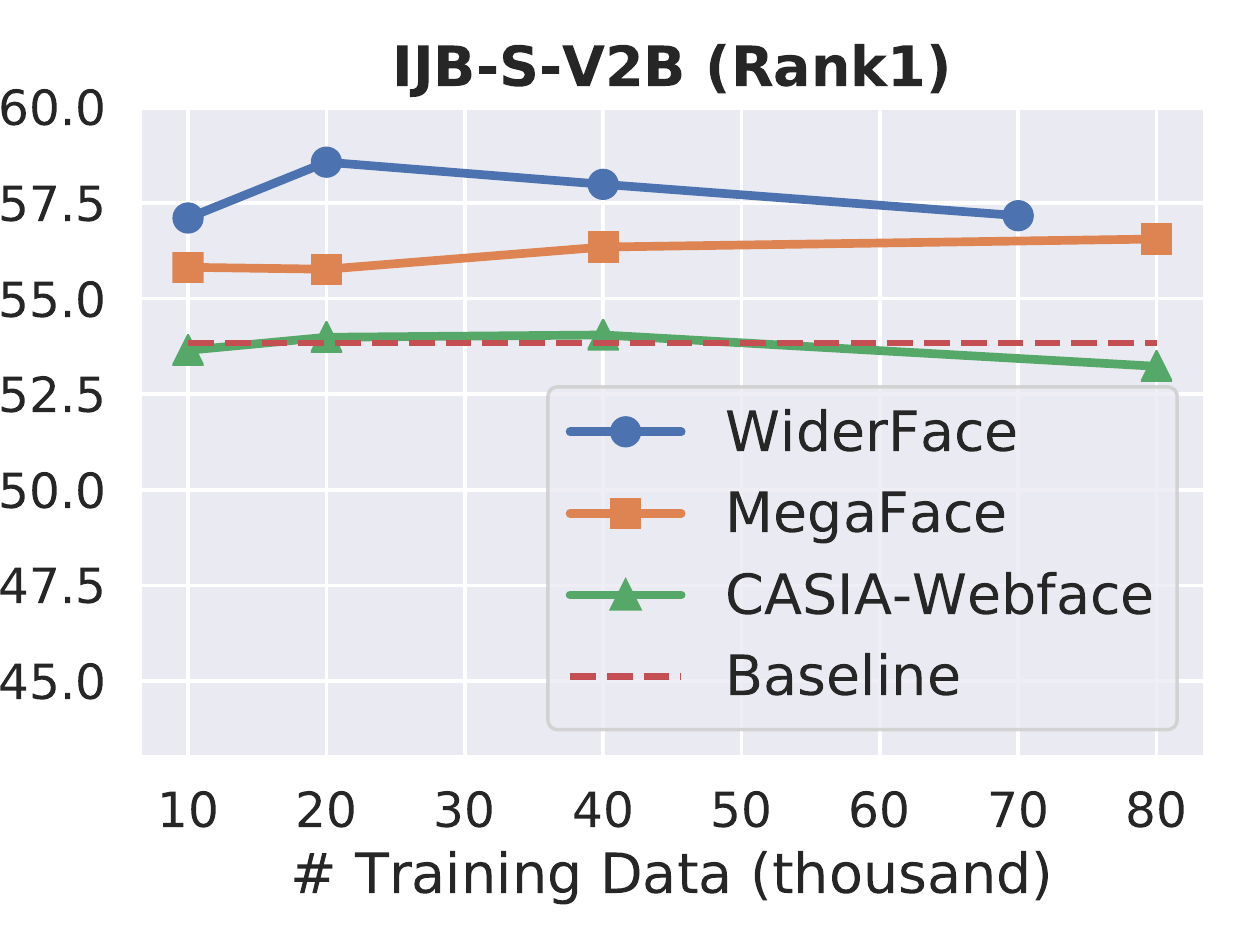}\\
    \includegraphics[width=0.49\linewidth]{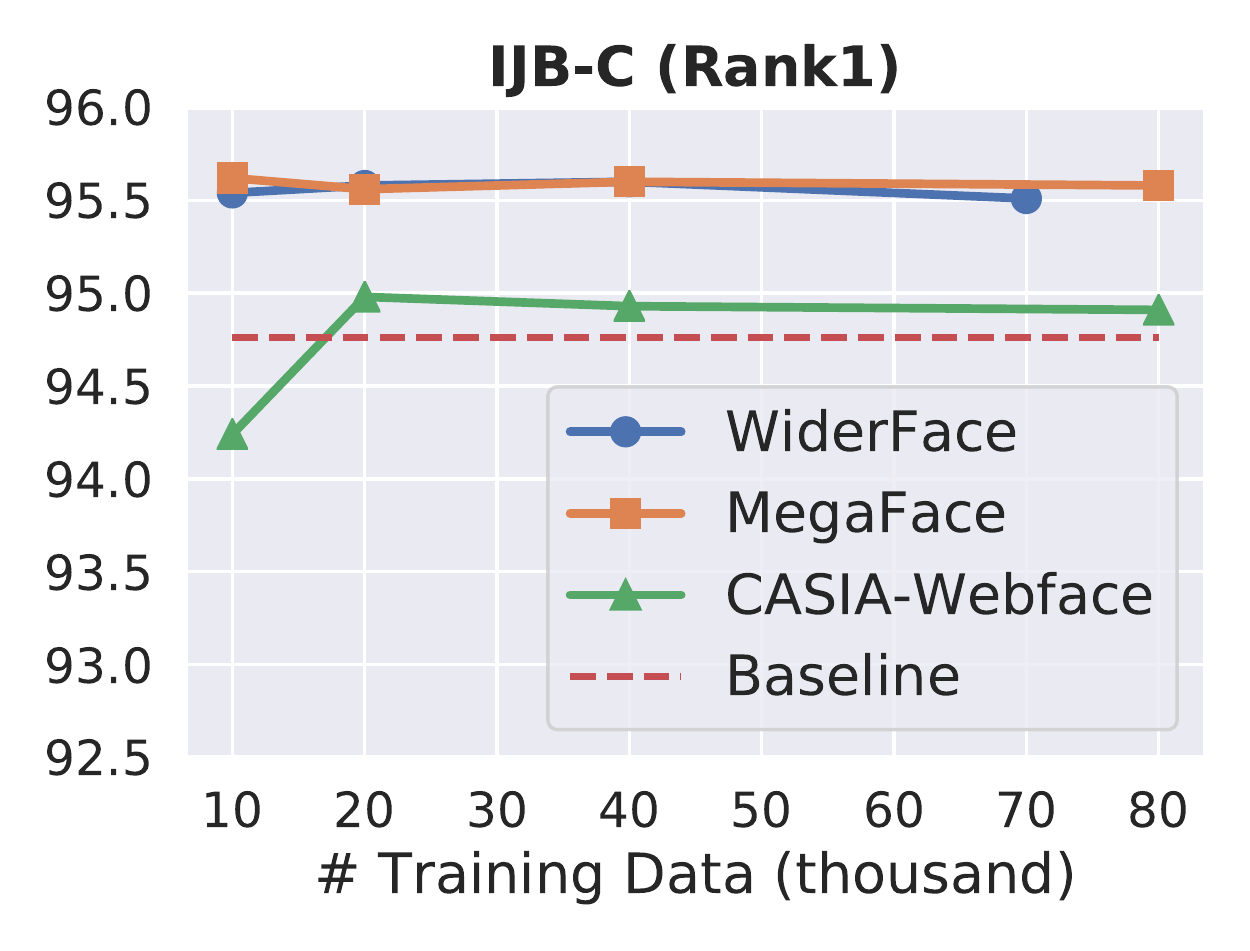}\hfill
    \includegraphics[width=0.49\linewidth]{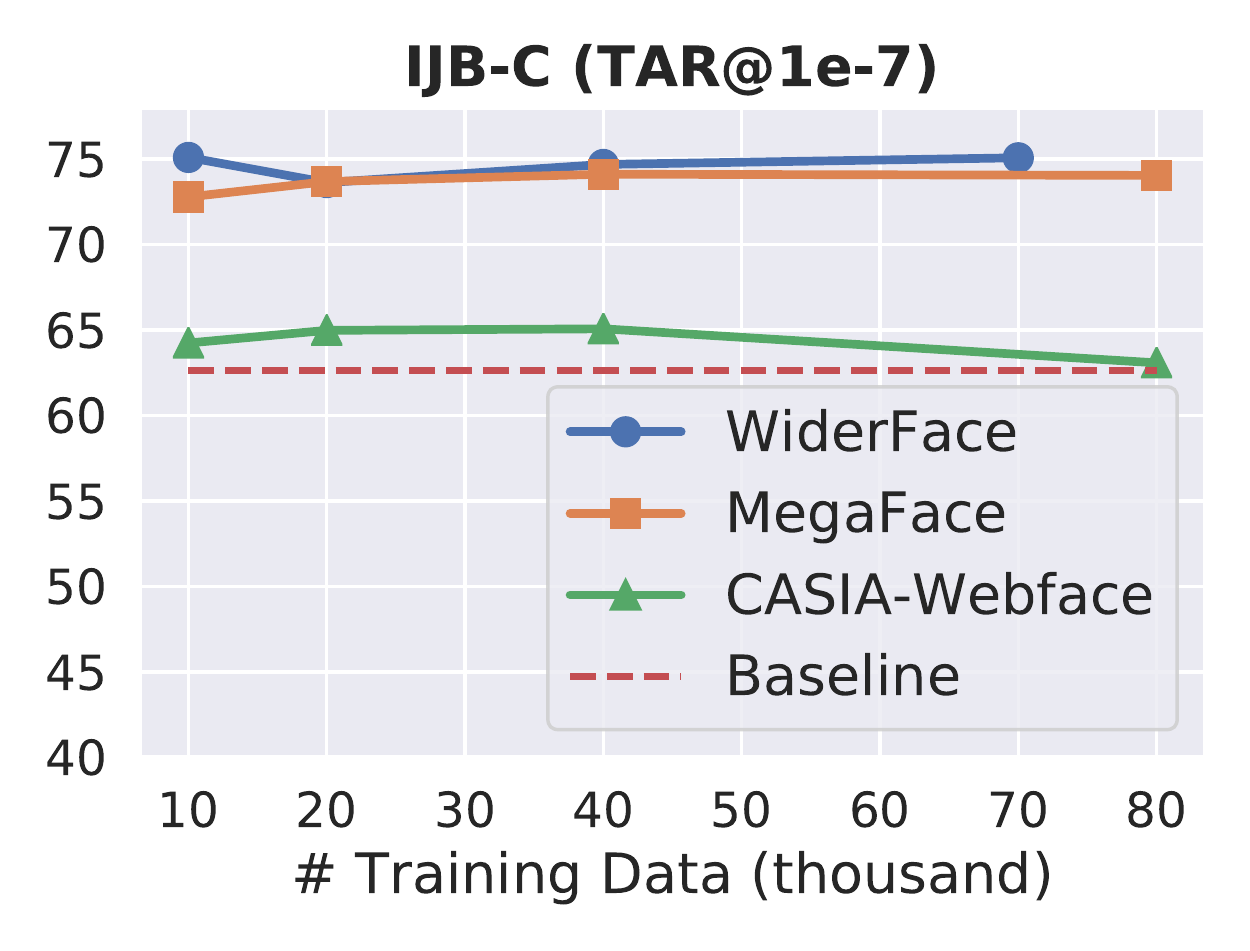}\\[-0.5em]
    \caption{Evaluation Results on IJB-S and IJB-C with different protocols and different number and choice of unlabeled training data. The red line here refers the performance of the supervised baseline which does not use any unlabeled data.}\vspace{-1.0em}
    \label{fig:appendix_unlabeled}
\end{figure}

\section{Choice of the Unlabeled Dataset}
\label{sec:appendix_unlabeled}

In Section~\ref{sec:exp_quantity}, we discussed on the impact of the quantity/diversity of training data on feature generalizability. A remaining question is how the choice and amount of unlabeled data would affect the performance. Here, we show additional experiments on different choices of unlabeled dataset. In addition to the WiderFace, we consider to utilize two other datasets: MegaFace~\cite{MegaFace} and CASIA-WebFace~\cite{yi2014learning}. For MegaFace, we only use the distractor images in their identification protocol, which are crawled from album photos on Flicker and present a larger degree of variation compared with the faces in MS-Celeb-1M. CASIA-WebFace, similar to MS-Celeb-1M, is mainly composed of celebrity photos, and therefore it should not introduce much additional diversity. Note that CASIA-WebFace is a labeled dataset but we ignore its labels for this experiment. The diversity (facial variation) of the three datasets can be ranked as: \textit{WiderFace $>$ MegaFace $>$ CASIA-WebFace}. Example images of the three datasets are shown in Figure~\ref{fig:exp_dataset}. For both MegaFace and CASIA-Webface, we choose a random subset to match the number of the WiderFace. Furthermore, to see the impact of the quantity of unlabeled dataset, we also train the models with different numbers of unlabeled data. Then, we evaluate all the models on IJB-S, IJB-C and LFW. The reason to evaluate on LFW here is to see the impact of different unlabeled datasets on the performance in the original domain. The results are shown in Fig.~\ref{fig:appendix_unlabeled}. Note that due to the large number of experiments, we do not use augmentation network here. But empirically we found the trends are similar with the data augmentation network.

From Fig.~\ref{fig:appendix_unlabeled}, it can be seen that in general, the more diverse the unlabeled dataset is, the more performance boost it leads to. In particular, using CASIA-WebFace as the unlabeled dataset hardly improves performance on any protocol. This is expected because CASIA-WebFace is very similar to MS-Celeb-1M and hence it cannot introduce additional diversity to regularize the training of face representations. Using MegaFace distractors as the unlabeled dataset improves the performance on both IJB-C and IJB-S, both of which have more variations than the MS-Celeb-1M. Using WiderFace as the unlabeled dataset further improves the performance on the IJB-S dataset. Note that all the models in this experiment maintain the high performance on the LFW dataset. In other words, \textit{using a more diverse unlabeled dataset would not impair the performance on the original domain and safely improves the performance on the challenging new domains}. An additional result that we can observe is that the size of the unlabeled dataset does not have a clear effect compared to its diversity.

\begin{table}[t]
\captionsetup{font=small}
\newcommand{\mr}[1]{\multirow{2}{*}{#1}}
\setlength{\tabcolsep}{1.2pt}
\begin{center}
\scriptsize
\begin{tabularx}{1.0\linewidth}{X | c|c || c|c|c|c |c|c}
\Xhline{2\arrayrulewidth}
\mr{Method}                  & \mr{Data} & \mr{Model} & \multicolumn{4}{c|}{Verification} & \multicolumn{2}{c}{Identification} \\\cline{4-9}                            &  & & 1e-7 & 1e-6 & 1e-5 & 1e-4 & Rank1 & Rank5 \\
\Xhline{2\arrayrulewidth}
Cao et al.~\cite{cao2018vggface2} & 13.3M   & SE-ResNet-50 & - & - & 76.8 & 86.2 & 91.4 & 95.1 \\\hline
PFE~\cite{shi2019probabilistic} & 4.4M      & ResNet-64 & - & - & 89.64 & 93.25 & 95.49 &  97.17 \\\hline
ArcFace~\cite{deng2018arcface}              & 5.8M & ResNet-50 & 67.40 & 80.52 & 88.36 & 92.52 & 93.26 & 95.33 \\\hline
Ranjan et al.~\cite{ranjan2019fast}         & 5.6M & ResNet-101 & 67.4 & 76.4 & 86.2 & 91.9 & 94.6 & 97.5 \\\hline
AFRN~\cite{kang2019attentional}             & 3.1M & ResNet-101 & - & - & 88.3 & 93.0 & \textbf{95.7} & \textbf{97.6} \\\hline
DUL~\cite{chang2020data}                    & 3.6M & ResNet-64 & - & - & 90.23 & 94.2 & \textbf{95.7} & \textbf{97.6} \\\hline
\Xhline{2\arrayrulewidth}
Baseline                    & 3.9M & ResNet-50 & 62.90 & 82.94 & 90.73 & 94.57 & 94.90 & 96.77 \\\hline
Proposed                    & 4.0M & ResNet-50 & \textbf{77.39} & \textbf{87.92} & \textbf{91.86} & \textbf{94.66} & 95.61 & 97.13 \\\hline
\Xhline{2\arrayrulewidth}
\end{tabularx}\vspace{-1.0em}
\caption{Performance comparison with state-of-the-art methods on the IJB-C dataset.}\vspace{-1.0em}
\label{tab:ijbc}
\end{center}
\end{table}

\begin{table}[t]
\captionsetup{font=small}
\newcommand{\mr}[1]{\multirow{2}{*}{#1}}
\setlength{\tabcolsep}{1.2pt}
\begin{center}
\scriptsize
\begin{tabularx}{1.00\linewidth}{X | c|c || c|c|c|c|c|c}
\Xhline{2\arrayrulewidth}
\mr{Method}                  & \mr{Data} & \mr{Model} & \multicolumn{4}{c|}{Verification} & \multicolumn{2}{c}{Identification} \\\cline{4-9}
                            &  & & 1e-6 & 1e-5 & 1e-4 & 1e-3 & Rank1 & Rank5 \\
\Xhline{2\arrayrulewidth}
Cao et al.~\cite{cao2018vggface2}   & 13.3M & SE-ResNet-50 & - & 70.5 & 83.1 & 90.8 & 90.2 & 94.6 \\\hline
Comparator~\cite{xie2018comparator} & 3.3M & ResNet-50 & - & - & 84.9 & 93.7 & - & - \\\hline
ArcFace~\cite{deng2018arcface}      & 5.8M & ResNet-50 & 40.77 & 84.28 & 91.66 & 94.81 & 92.95 & 95.60 \\\hline
Ranjan et al.~\cite{ranjan2019fast} & 5.6M & ResNet-101 & \textbf{48.4} & 80.4 & 89.8 & 94.4 & 93.3 & 96.6 \\\hline
AFRN~\cite{kang2019attentional}     & 3.1M & ResNet-101 & - & 77.1 & 88.5 & 94.9 & \textbf{97.3} & \textbf{97.6} \\\hline
\Xhline{2\arrayrulewidth}
Baseline                    & 3.9M & ResNet-50 & 40.12 & 84.38 & \textbf{92.79} & \textbf{95.90} & 93.85 & 96.55 \\\hline
Proposed                    & 4.0M & ResNet-50 & 43.38 & \textbf{88.19} & 92.78 & 95.86 & 94.62 & 96.72 \\\hline
\Xhline{2\arrayrulewidth}
\end{tabularx}\vspace{-1.0em}
\caption{Performance comparison with state-of-the-art methods on the IJB-B dataset.}\vspace{-1.0em}
\label{tab:ijbb}
\end{center}
\end{table}

\subsection{Comparison with State-of-the-Art FR Methods}
\label{sec:exp_final}
In Table~\ref{tab:ijbc} we show more complete results on IJB-C dataset and compare our method with other state-of-the-art methods. In generally, we observe that with fewer labeled training samples and number of parameters, we are able to achieve state-of-the-art performance on most of the protocols. Particularly at low FARs, the proposed method outperforms the baseline methods with a good margin. This is because at a low FAR, the verification threshold is mainly determined by low quality impostor pairs, which are instances of the difficult face samples that we are targeting with additional unlabeled data. Similar trend is observed for IJB-B dataset (Table~\ref{tab:ijbb}). Note that because of fewer number of face pairs, we are only able to test at higher FARs for IJB-B dataset.

In Table~\ref{tab:ijbs} we show the results on two different protocols of IJB-S. Both the Surveillance-to-Still (V2S) and Surveillance-to-Booking (V2B) protocols use surveillance videos as probes and mugshots as gallery. Therefore, IJB-S results represent a cross domain comparison problem. Overall, the proposed system achieve new state-of-the-art performance on both protocols.

\begin{table}[t]
\captionsetup{font=small}
\setlength{\tabcolsep}{0.8pt}
\begin{center}
\scriptsize
\begin{tabularx}{1.0\linewidth}{X || c|c|c|c|c|| c|c|c|c|c}
\Xhline{2\arrayrulewidth}
\multirow{2}{*}{Method}      & \multicolumn{5}{c||}{Surveillance-to-Still} & \multicolumn{5}{c}{Surveillance-to-Booking} \\\cline{2-11}
                            & Rank1 & Rank5 & Rank10 & 1\% & 10\% & Rank1 & Rank5 & Rank10 & 1\% & 10\% \\
\Xhline{2\arrayrulewidth}
MARN~\cite{gong2019low}             & 58.14 & 64.11 & - & 21.47 & - & 59.26 & 65.93 & - & 32.07 & - \\\hline
PFE~\cite{shi2019probabilistic}     & 50.16 & 58.33 & 62.28 & 31.88 & 35.33 & 53.60 & 61.75 & 62.97 & 35.99 & 39.82 \\\hline
ArcFace\cite{deng2018arcface}       & 50.39 & 60.42 & 64.74 & 32.39 & 42.99 & 52.25 & 61.19 & 65.63 & 34.87 & 43.50 \\
\Xhline{2\arrayrulewidth}
Baseline                            & 53.23 & 62.91 & 67.83 & 31.88 & 43.32 & 54.26 & 64.18 & 69.26 & 32.39 & 44.32 \\\hline
Proposed                            & \textbf{59.29} & \textbf{66.91} & \textbf{69.63} & \textbf{39.92} & \textbf{50.49} & \textbf{60.58} & \textbf{67.70} & \textbf{70.63} & \textbf{40.80} & \textbf{50.31} \\\hline
\Xhline{2\arrayrulewidth}
\end{tabularx}\vspace{-0.8em}
\caption{Performance on the IJB-S benchmark.}\vspace{-1.0em}
\label{tab:ijbs}
\end{center}
\end{table}

\section{Conclusions}
We have proposed a semi-supervised framework of learning robust face representation that could generalize to unconstrained faces beyond the labeled training data. Without collecting domain specific data, we utilized a relatively small unlabeled dataset containing diverse styles of face images. In order to fully utilize the unlabeled dataset, two methods are proposed. First, we showed that the domain adversarial learning, which is common in adaptation methods, can be applied in our setting to reduce domain gaps between labeled faces and hidden sub-domains. Second, we propose an augmentation network that can capture different visual styles in the unlabeled dataset and apply them to the labeled images during training, making the face representation more discriminative for unconstrained faces. Our experimental results show that as the number of labeled images increases, the performance of the supervised baseline tends to saturate on the challenging testing scenarios. Instead, introducing more diverse training data becomes more important and helpful. In a few challenging protocols, we showed that the proposed method can outperform the supervised baseline with less than half of the labeled data. By training on the labeled MS-Celeb-1M dataset and unlabeled WiderFace dataset, our final model achieves state-of-the-art performance on challenging benchmarks such as IJB-B, IJB-C and IJB-S.

{\small
\bibliographystyle{ieee_fullname}
\bibliography{egbib}

\begin{thebibliography}{10}\itemsep=-1pt

\bibitem{LayerNorm}
Jimmy~Lei Ba, Jamie~Ryan Kiros, and Geoffrey~E Hinton.
\newblock Layer normalization.
\newblock {\em arXiv:1607.06450}, 2016.

\bibitem{berthelot2019mixmatch}
David Berthelot, Nicholas Carlini, Ian Goodfellow, Nicolas Papernot, Avital
  Oliver, and Colin~A Raffel.
\newblock Mixmatch: A holistic approach to semi-supervised learning.
\newblock In {\em NeurIPS}, 2019.

\bibitem{cao2018vggface2}
Qiong Cao, Li Shen, Weidi Xie, Omkar~M Parkhi, and Andrew Zisserman.
\newblock Vggface2: A dataset for recognising faces across pose and age.
\newblock In {\em IEEE FG}, 2018.

\bibitem{carlucci2019domain}
Fabio~M Carlucci, Antonio D'Innocente, Silvia Bucci, Barbara Caputo, and
  Tatiana Tommasi.
\newblock Domain generalization by solving jigsaw puzzles.
\newblock In {\em CVPR}, 2019.

\bibitem{chang2020data}
Jie Chang, Zhonghao Lan, Changmao Cheng, and Yichen Wei.
\newblock Data uncertainty learning in face recognition.
\newblock In {\em CVPR}, 2020.

\bibitem{deng2018arcface}
Jiankang Deng, Jia Guo, and Stefanos Zafeiriou.
\newblock Arcface: Additive angular margin loss for deep face recognition.
\newblock {\em CVPR}, 2019.

\bibitem{deng2019retinaface}
Jiankang Deng, Jia Guo, Yuxiang Zhou, Jinke Yu, Irene Kotsia, and Stefanos
  Zafeiriou.
\newblock Retinaface: Single-stage dense face localisation in the wild.
\newblock {\em arXiv preprint arXiv:1905.00641}, 2019.

\bibitem{ganin2014unsupervised}
Yaroslav Ganin and Victor Lempitsky.
\newblock Unsupervised domain adaptation by backpropagation.
\newblock 2015.

\bibitem{ghifary2015domain}
Muhammad Ghifary, W Bastiaan~Kleijn, Mengjie Zhang, and David Balduzzi.
\newblock Domain generalization for object recognition with multi-task
  autoencoders.
\newblock In {\em ICCV}, 2015.

\bibitem{gong2019low}
Sixue Gong, Yichun Shi, and Anil Jain.
\newblock Low quality video face recognition: Multi-mode aggregation recurrent
  network (marn).
\newblock In {\em ICCV Workshops}, 2019.

\bibitem{guo2020learning}
Jianzhu Guo, Xiangyu Zhu, Chenxu Zhao, Dong Cao, Zhen Lei, and Stan~Z Li.
\newblock Learning meta face recognition in unseen domains.
\newblock In {\em CVPR}, 2020.

\bibitem{guo2016msceleb}
Yandong Guo, Lei Zhang, Yuxiao Hu, Xiaodong He, and Jianfeng Gao.
\newblock Ms-celeb-1m: A dataset and benchmark for large scale face
  recognition.
\newblock In {\em ECCV}, 2016.

\bibitem{hasnat2017deepvisage}
Abul Hasnat, Julien Bohn{\'e}, Jonathan Milgram, St{\'e}phane Gentric, and
  Liming Chen.
\newblock Deepvisage: Making face recognition simple yet with powerful
  generalization skills.
\newblock {\em ICCV}, 2017.

\bibitem{hoffman2018cycada}
Judy Hoffman, Eric Tzeng, Taesung Park, Jun-Yan Zhu, Phillip Isola, Kate
  Saenko, Alexei Efros, and Trevor Darrell.
\newblock Cycada: Cycle-consistent adversarial domain adaptation.
\newblock In {\em ICML}, 2018.

\bibitem{LFWTech}
Gary~B. Huang, Manu Ramesh, Tamara Berg, and Erik Learned-Miller.
\newblock Labeled faces in the wild: A database for studying face recognition
  in unconstrained environments.
\newblock Technical Report 07-49, University of Massachusetts, Amherst, October
  2007.

\bibitem{huang2017adain}
Xun Huang and Serge~J Belongie.
\newblock Arbitrary style transfer in real-time with adaptive instance
  normalization.
\newblock In {\em ICCV}, 2017.

\bibitem{MUNIT}
Xun Huang, Ming-Yu Liu, Serge Belongie, and Jan Kautz.
\newblock Multimodal unsupervised image-to-image translation.
\newblock In {\em ECCV}, 2018.

\bibitem{IJBS}
Nathan~D. Kalka, Brianna Maze, James~A. Duncan, Kevin~J. O’Connor, Stephen
  Elliott, Kaleb Hebert, Julia Bryan, and Anil~K. Jain.
\newblock {IJB-S : IARPA Janus Surveillance Video Benchmark }.
\newblock In {\em BTAS}, 2018.

\bibitem{kang2019attentional}
Bong-Nam Kang, Yonghyun Kim, Bongjin Jun, and Daijin Kim.
\newblock Attentional feature-pair relation networks for accurate face
  recognition.
\newblock In {\em ICCV}, 2019.

\bibitem{kang2019contrastive}
Guoliang Kang, Lu Jiang, Yi Yang, and Alexander~G Hauptmann.
\newblock Contrastive adaptation network for unsupervised domain adaptation.
\newblock In {\em CVPR}, 2019.

\bibitem{MegaFace}
Ira Kemelmacher-Shlizerman, Steven~M Seitz, Daniel Miller, and Evan Brossard.
\newblock The megaface benchmark: 1 million faces for recognition at scale.
\newblock In {\em CVPR}, 2016.

\bibitem{IJBA}
Brendan~F Klare, Ben Klein, Emma Taborsky, Austin Blanton, Jordan Cheney,
  Kristen Allen, Patrick Grother, Alan Mah, and Anil~K Jain.
\newblock Pushing the frontiers of unconstrained face detection and
  recognition: {IARPA Janus Benchmark A}.
\newblock In {\em CVPR}, 2015.

\bibitem{laine2017temporal}
Samuli Laine and Timo Aila.
\newblock Temporal ensembling for semi-supervised learning.
\newblock 2017.

\bibitem{lee2013pseudo}
Dong-Hyun Lee.
\newblock Pseudo-label: The simple and efficient semi-supervised learning
  method for deep neural networks.
\newblock In {\em ICML Workshop}, 2013.

\bibitem{li2018domain}
Haoliang Li, Sinno Jialin~Pan, Shiqi Wang, and Alex~C Kot.
\newblock Domain generalization with adversarial feature learning.
\newblock In {\em CVPR}, 2018.

\bibitem{liu2017sphereface}
Weiyang Liu, Yandong Wen, Zhiding Yu, Ming Li, Bhiksha Raj, and Le Song.
\newblock Sphereface: Deep hypersphere embedding for face recognition.
\newblock In {\em CVPR}, 2017.

\bibitem{long2017deep}
Mingsheng Long, Han Zhu, Jianmin Wang, and Michael~I Jordan.
\newblock Deep transfer learning with joint adaptation networks.
\newblock In {\em ICML}, 2017.

\bibitem{masi2016we}
Iacopo Masi, Anh~Tuan Tran, Tal Hassner, Jatuporn~Toy Leksut, and G{\'e}rard
  Medioni.
\newblock Do we really need to collect millions of faces for effective face
  recognition?
\newblock In {\em ECCV}, 2016.

\bibitem{IJBC}
Brianna Maze, Jocelyn Adams, James~A Duncan, Nathan Kalka, Tim Miller, Charles
  Otto, Anil~K Jain, W~Tyler Niggel, Janet Anderson, Jordan Cheney, et~al.
\newblock Iarpa janus benchmark-c: Face dataset and protocol.
\newblock In {\em ICB}, 2018.

\bibitem{motiian2017unified}
Saeid Motiian, Marco Piccirilli, Donald~A Adjeroh, and Gianfranco Doretto.
\newblock Unified deep supervised domain adaptation and generalization.
\newblock In {\em ICCV}, 2017.

\bibitem{muandet2013domain}
Krikamol Muandet, David Balduzzi, and Bernhard Sch{\"o}lkopf.
\newblock Domain generalization via invariant feature representation.
\newblock In {\em ICML}, 2013.

\bibitem{pan2010domain}
Sinno~Jialin Pan, Ivor~W Tsang, James~T Kwok, and Qiang Yang.
\newblock Domain adaptation via transfer component analysis.
\newblock {\em IEEE Trans. on Neural Networks}, 2010.

\bibitem{ranjan2019fast}
Rajeev Ranjan, Ankan Bansal, Jingxiao Zheng, Hongyu Xu, Joshua Gleason, Boyu
  Lu, Anirudh Nanduri, Jun-Cheng Chen, Carlos~D Castillo, and Rama Chellappa.
\newblock A fast and accurate system for face detection, identification, and
  verification.
\newblock {\em IEEE Trans. on Biometrics, Behavior, and Identity Science},
  2019.

\bibitem{ranjan2017l2}
Rajeev Ranjan, Carlos~D Castillo, and Rama Chellappa.
\newblock L2-constrained softmax loss for discriminative face verification.
\newblock {\em arXiv:1703.09507}, 2017.

\bibitem{rasmus2015semi}
Antti Rasmus, Mathias Berglund, Mikko Honkala, Harri Valpola, and Tapani Raiko.
\newblock Semi-supervised learning with ladder networks.
\newblock In {\em NeurIPS}, 2015.

\bibitem{saito2018maximum}
Kuniaki Saito, Kohei Watanabe, Yoshitaka Ushiku, and Tatsuya Harada.
\newblock Maximum classifier discrepancy for unsupervised domain adaptation.
\newblock In {\em CVPR}, 2018.

\bibitem{schroff2015facenet}
Florian Schroff, Dmitry Kalenichenko, and James Philbin.
\newblock Facenet: A unified embedding for face recognition and clustering.
\newblock In {\em CVPR}, 2015.

\bibitem{shi2019probabilistic}
Yichun Shi and Anil~K Jain.
\newblock Probabilistic face embeddings.
\newblock In {\em ICCV}, 2019.

\bibitem{sohn2016improved}
Kihyuk Sohn.
\newblock Improved deep metric learning with multi-class n-pair loss objective.
\newblock In {\em NIPS}, 2016.

\bibitem{sohn2020fixmatch}
Kihyuk Sohn, David Berthelot, Chun-Liang Li, Zizhao Zhang, Nicholas Carlini,
  Ekin~D Cubuk, Alex Kurakin, Han Zhang, and Colin Raffel.
\newblock Fixmatch: Simplifying semi-supervised learning with consistency and
  confidence.
\newblock {\em arXiv:2001.07685}, 2020.

\bibitem{sohn2017unsupervised}
Kihyuk Sohn, Wendy Shang, Xiang Yu, and Manmohan Chandraker.
\newblock Unsupervised domain adaptation for face recognition in unlabeled
  videos.
\newblock In {\em CVPR}, 2017.

\bibitem{deepid2}
Yi Sun, Yuheng Chen, Xiaogang Wang, and Xiaoou Tang.
\newblock Deep learning face representation by joint
  identification-verification.
\newblock In {\em NIPS}, 2014.

\bibitem{sun2020circle}
Yifan Sun, Changmao Cheng, Yuhan Zhang, Chi Zhang, Liang Zheng, Zhongdao Wang,
  and Yichen Wei.
\newblock Circle loss: A unified perspective of pair similarity optimization.
\newblock {\em arXiv:2002.10857}, 2020.

\bibitem{taigman2014deepface}
Yaniv Taigman, Ming Yang, Marc'Aurelio Ranzato, and Lior Wolf.
\newblock Deepface: Closing the gap to human-level performance in face
  verification.
\newblock In {\em CVPR}, 2014.

\bibitem{tarvainen2017mean}
Antti Tarvainen and Harri Valpola.
\newblock Mean teachers are better role models: Weight-averaged consistency
  targets improve semi-supervised deep learning results.
\newblock In {\em NeurIPS}, 2017.

\bibitem{tzeng2017adversarial}
Eric Tzeng, Judy Hoffman, Kate Saenko, and Trevor Darrell.
\newblock Adversarial discriminative domain adaptation.
\newblock In {\em CVPR}, 2017.

\bibitem{InstanceNorm}
Dmitry Ulyanov, Andrea Vedaldi, and Victor Lempitsky.
\newblock Instance normalization: The missing ingredient for fast stylization.
\newblock {\em arXiv:1607.08022}, 2016.

\bibitem{wang2018additive}
Feng Wang, Weiyang Liu, Haijun Liu, and Jian Cheng.
\newblock Additive margin softmax for face verification.
\newblock {\em arXiv:1801.05599}, 2018.

\bibitem{wang2018cosface}
Hao Wang, Yitong Wang, Zheng Zhou, Xing Ji, Zhifeng Li, Dihong Gong, Jingchao
  Zhou, and Wei Liu.
\newblock Cosface: Large margin cosine loss for deep face recognition.
\newblock {\em CVPR}, 2018.

\bibitem{IJBB}
Cameron Whitelam, Emma Taborsky, Austin Blanton, Brianna Maze, Jocelyn Adams,
  Tim Miller, Nathan Kalka, Anil~K Jain, James~A Duncan, Kristen Allen, et~al.
\newblock Iarpa janus benchmark-b face dataset.
\newblock In {\em Proceedings of the IEEE Conference on Computer Vision and
  Pattern Recognition Workshops}, pages 90--98, 2017.

\bibitem{xie2019unsupervised}
Qizhe Xie, Zihang Dai, Eduard Hovy, Minh-Thang Luong, and Quoc~V Le.
\newblock Unsupervised data augmentation for consistency training.
\newblock {\em arXiv:1904.12848}, 2019.

\bibitem{xie2018comparator}
Weidi Xie, Li Shen, and Andrew Zisserman.
\newblock Comparator networks.
\newblock In {\em ECCv}, 2018.

\bibitem{yang2016wider}
Shuo Yang, Ping Luo, Chen~Change Loy, and Xiaoou Tang.
\newblock Wider face: A face detection benchmark.
\newblock In {\em CVPR}, 2016.

\bibitem{yi2014learning}
Dong Yi, Zhen Lei, Shengcai Liao, and Stan~Z Li.
\newblock Learning face representation from scratch.
\newblock {\em arXiv:1411.7923}, 2014.

\bibitem{zhai2019s4l}
Xiaohua Zhai, Avital Oliver, Alexander Kolesnikov, and Lucas Beyer.
\newblock S4l: Self-supervised semi-supervised learning.
\newblock In {\em ICCV}, 2019.

\bibitem{zhang2019adacos}
Xiao Zhang, Rui Zhao, Yu Qiao, Xiaogang Wang, and Hongsheng Li.
\newblock Adacos: Adaptively scaling cosine logits for effectively learning
  deep face representations.
\newblock In {\em CVPR}, 2019.

\bibitem{zhu2017unpaired}
Jun-Yan Zhu, Taesung Park, Phillip Isola, and Alexei~A Efros.
\newblock Unpaired image-to-image translation using cycle-consistent
  adversarial networks.
\newblock In {\em ICCV}, 2017.

\end{thebibliography}
}

\clearpage
\section*{Appendix}

\appendix

\section{Numerical Results on IJB-B, IJB-C}
In Table~\ref{tab:appendix_ijbc} and Table~\ref{tab:appendix_ijbb}, we show more numerical results on the IJB-C and IJB-B dataset, respectively. Since all the baseline methods (from other papers) are trained on different number of labeled images, we report the performance of our models trained on different labeled subsets
for a more fair comparison. From the tables, we could observe that our models outperform most of the baselines with equal or less than 2M labeled data.

\begin{table}[h]
\captionsetup{font=small}
\newcommand{\mr}[1]{\multirow{2}{*}{#1}}
\setlength{\tabcolsep}{0.6pt}
\begin{center}
\scriptsize
\begin{tabularx}{1.0\linewidth}{X | c|c || c|c|c|c|c|c}
\Xhline{2\arrayrulewidth}
\mr{Method}                  & \mr{Data} & \mr{Model} & \multicolumn{4}{c|}{Verification} & \multicolumn{2}{c}{Identification} \\\cline{4-9}                            &  & & 1e-7 & 1e-6 & 1e-5 & 1e-4 & Rank1 & Rank5 \\
\Xhline{2\arrayrulewidth}
Cao et al.~\cite{cao2018vggface2} & 13.3M & SE-ResNet-50 & - & - & 76.8 & 86.2 & 91.4 & 95.1 \\\hline
PFE~\cite{shi2019probabilistic} & 4.4M & ResNet-64 & - & - & 89.64 & 93.25 & 95.49 &  97.17 \\\hline
ArcFace~\cite{deng2018arcface}          & 5.8M & ResNet-50 & 67.40 & 80.52 & 88.36 & 92.52 & 93.26 & 95.33 \\\hline
Ranjan et al.~\cite{ranjan2019fast} & 5.6M & ResNet-101 & 67.4 & 76.4 & 86.2 & 91.9 & 94.6 & 97.5 \\\hline
AFRN~\cite{kang2019attentional}     & 3.1M & ResNet-101 & - & - & 88.3 & 93.0 & \textbf{95.7} & \textbf{97.6} \\\hline
\Xhline{2\arrayrulewidth}
Baseline                    & 500K & ResNet-50      & 51.13 & 66.44 & 77.58 & 87.73 & 90.90 & 94.50 \\
Proposed                    & 500K+70K & ResNet-50  & 60.33 & 71.24 & 80.31 & 88.18 & 91.81 & 94.96 \\\hline
Baseline                    & 1.0M & ResNet-50      & 59.53 & 77.70 & 86.16 & 92.13 & 93.62 & 95.93 \\
Proposed                    & 1.0M+70K & ResNet-50  & 61.87 & 79.76 & 87.16 & 92.39 & 94.19 & 96.30 \\\hline
Baseline                    & 2.0M & ResNet-50      & 67.64 & 78.66 & 88.16 & 93.48 & 94.34 & 96.34 \\
Proposed                    & 2.0M+70K & ResNet-50  & \textbf{78.62} & 84.91 & 90.61 & 93.77 & 95.04 & 96.80 \\\hline
Baseline                    & 3.0M & ResNet-50      & 62.65 & 79.20 & 89.20 & 94.20 & 94.76 & 96.49 \\
Proposed                    & 3.0M+70K & ResNet-50  & 78.38 & 85.91 & 91.56 & 94.48 & 95.51 & 97.04 \\\hline
Baseline                    & 3.9M & ResNet-50      & 62.90 & 82.94 & 90.73 & 94.57 & 94.90 & 96.77 \\
Proposed                    & 3.9M+70K & ResNet-50  & 77.39 & \textbf{87.92} & \textbf{91.86} & \textbf{94.66} & 95.61 & 97.13 \\\hline
\Xhline{2\arrayrulewidth}
\end{tabularx}\vspace{-2em}
\end{center}
\caption{Performance comparison with state-of-the-art methods on the IJB-C dataset.}\vspace{-0.5em}
\label{tab:appendix_ijbc}
\end{table}

\begin{table}[h]
\captionsetup{font=small}
\newcommand{\mr}[1]{\multirow{2}{*}{#1}}
\setlength{\tabcolsep}{0.6pt}
\begin{center}
\scriptsize
\begin{tabularx}{1.00\linewidth}{X | c|c || c|c|c|c|c|c}
\Xhline{2\arrayrulewidth}
\mr{Method}                  & \mr{Data} & \mr{Model} & \multicolumn{4}{c|}{Verification} & \multicolumn{2}{c}{Identification} \\\cline{4-9}
                            &  & & 1e-6 & 1e-5 & 1e-4 & 1e-3 & Rank1 & Rank5 \\
\Xhline{2\arrayrulewidth}
Cao et al.~\cite{cao2018vggface2}   & 13.3M & SE-ResNet-50 & - & 70.5 & 83.1 & 90.8 & 90.2 & 94.6 \\\hline
Comparator~\cite{xie2018comparator} & 3.3M & ResNet-50 & - & - & 84.9 & 93.7 & - & - \\\hline
ArcFace~\cite{deng2018arcface}      & 5.8M & ResNet-50 & 40.77 & 84.28 & 91.66 & 94.81 & 92.95 & 95.60 \\\hline
Ranjan et al.~\cite{ranjan2019fast} & 5.6M & ResNet-101 & \textbf{48.4} & 80.4 & 89.8 & 94.4 & 93.3 & 96.6 \\\hline
AFRN~\cite{kang2019attentional}     & 3.1M & ResNet-101 & - & 77.1 & 88.5 & 94.9 & \textbf{97.3} & \textbf{97.6} \\\hline
\Xhline{2\arrayrulewidth}
Baseline                    & 500K & ResNet-50      & 39.35 & 71.14 & 84.37 & 92.12 & 89.74 & 94.16 \\
Proposed                    & 500K+70K & ResNet-50  & 45.39 & 72.35 & 84.75 & 92.00 & 90.46 & 94.42 \\\hline
Baseline                    & 1.0M & ResNet-50      & 45.75 & 80.11 & 90.19 & 94.48 & 92.37 & 95.78 \\
Proposed                    & 1.0M+70K & ResNet-50  & 41.59 & 82.10 & 90.09 & 94.64 & 92.88 & 95.91 \\\hline
Baseline                    & 2.0M & ResNet-50      & 47.62 & 82.30 & 91.82 & 95.46 & 93.25 & 96.05 \\
Proposed                    & 2.0M+70K & ResNet-50  & 44.76 & 86.26 & 91.92 & 95.27 & 94.01 & 96.23 \\\hline
Baseline                    & 3.0M & ResNet-50      & 42.77 & 82.86 & 92.48 & 95.78 & 93.80 & 96.23 \\
Proposed                    & 3.0M+70K & ResNet-50  & 43.09 & 87.31 & \textbf{92.80} & 95.70 & 94.35 & 96.53 \\\hline
Baseline                    & 3.9M & ResNet-50      & 40.12 & 84.38 & 92.79 & \textbf{95.90} & 93.85 & 96.55 \\
Proposed                    & 3.9M+70K & ResNet-50  & 43.38 & \textbf{88.19} & 92.78 & 95.86 & 94.62 & 96.72 \\\hline
\Xhline{2\arrayrulewidth}
\end{tabularx}\vspace{-2em}
\end{center}
\caption{Performance comparison with state-of-the-art methods on the IJB-B dataset.}\vspace{-0.5em}
\label{tab:appendix_ijbb}
\end{table}

\section{Architecture of Augmentation Network}
The architecture of our augmentation network is based on MUNIT~\cite{MUNIT}. Let \texttt{c5s1-k} be a $5\times5$ convolutional layer with $k$ filters and stride $1$. \texttt{dk-IN} denotes a $3\times 3$ convolutional layer with $k$ filters and dilation $2$, where IN means Instance Normalization~\cite{InstanceNorm}. Similarly, AdaIN means Adaptive Instance Normalization~\cite{huang2017adain} and LN denotes Layer Normalization~\cite{LayerNorm}. \texttt{fc8} denotes a fully connected layer with $8$ filters. \texttt{avgpool} denotes a global average pooling layer. No normalization is used in the style encoder. We use Leaky ReLU with slope 0.2 in the discriminator and ReLU activation everywhere else. The architectures of different modules are as follows:
\begin{itemize}\vspace{-0.5em}
    \item Style Encoder: \\ \texttt{c5s1-32,c3s2-64,c3s2-128,avgpool,fc8} \vspace{-0.5em}
    \item Generator: \\ \texttt{c5s1-32-IN,d32-IN,d32-AdaIN,d32-LN,}\\\texttt{d32-LN,c5s1-3} \vspace{-0.5em}
    \item Discriminator: \\ \texttt{c5s1-32,c3s2-64,c3s2-128}
\end{itemize}
The length of the latent style code is set to $8$. A style decoder (multi-layer perceptron) has two hidden fully connected layers of $128$ filters without normalization, which transforms the latent style code to the parameters of the AdaIN layer.

\begin{figure*}[h]
\captionsetup{font=footnotesize}
\setlength\tabcolsep{1px}
\newcolumntype{Y}{>{\centering\arraybackslash}X}
    \centering
    \begin{tabularx}{\linewidth}{YYYYYYY}
        Input & Model (a) & Model (b) & Model (c) & Model (d)  & Model (e) & Model (f)  \\
    \midrule
        \includegraphics[width=0.97\linewidth]{fig/generated/1_0.png} & 
        \includegraphics[width=0.97\linewidth]{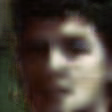} & \vspace{-0.97\linewidth}
        \includegraphics[width=0.48\linewidth]{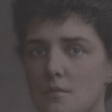} \hspace{-0.6em}
        \includegraphics[width=0.48\linewidth]{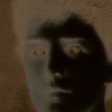} \hspace{-0.6em}
        \includegraphics[width=0.48\linewidth]{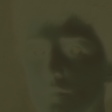} \hspace{-0.6em}
        \includegraphics[width=0.48\linewidth]{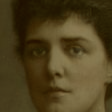} &  \vspace{-0.97\linewidth}
        \includegraphics[width=0.48\linewidth]{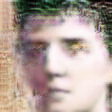} \hspace{-0.6em}
        \includegraphics[width=0.48\linewidth]{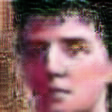} \hspace{-0.6em}
        \includegraphics[width=0.48\linewidth]{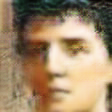} \hspace{-0.6em}
        \includegraphics[width=0.48\linewidth]{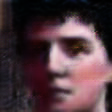} &  \vspace{-0.97\linewidth}
        \includegraphics[width=0.48\linewidth]{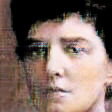} \hspace{-0.6em}
        \includegraphics[width=0.48\linewidth]{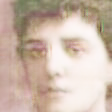} \hspace{-0.6em}
        \includegraphics[width=0.48\linewidth]{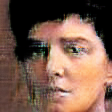} \hspace{-0.6em}
        \includegraphics[width=0.48\linewidth]{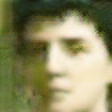} &  \vspace{-0.97\linewidth}
        \includegraphics[width=0.48\linewidth]{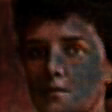} \hspace{-0.6em}
        \includegraphics[width=0.48\linewidth]{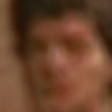} \hspace{-0.6em}
        \includegraphics[width=0.48\linewidth]{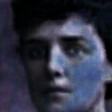} \hspace{-0.6em}
        \includegraphics[width=0.48\linewidth]{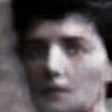} &  \vspace{-0.97\linewidth}
        \includegraphics[width=0.48\linewidth]{fig/generated/1_1.png} \hspace{-0.6em}
        \includegraphics[width=0.48\linewidth]{fig/generated/1_2.png} \hspace{-0.6em}
        \includegraphics[width=0.48\linewidth]{fig/generated/1_3.png} \hspace{-0.6em}
        \includegraphics[width=0.48\linewidth]{fig/generated/1_4.png} \\\midrule
        
        \includegraphics[width=0.97\linewidth]{fig/generated/2_0.png} & 
        \includegraphics[width=0.97\linewidth]{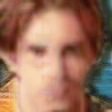} & \vspace{-0.97\linewidth}
        \includegraphics[width=0.48\linewidth]{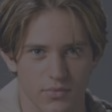} \hspace{-0.6em}
        \includegraphics[width=0.48\linewidth]{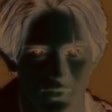} \hspace{-0.6em}
        \includegraphics[width=0.48\linewidth]{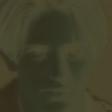} \hspace{-0.6em}
        \includegraphics[width=0.48\linewidth]{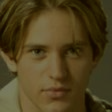} &  \vspace{-0.97\linewidth}
        \includegraphics[width=0.48\linewidth]{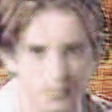} \hspace{-0.6em}
        \includegraphics[width=0.48\linewidth]{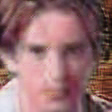} \hspace{-0.6em}
        \includegraphics[width=0.48\linewidth]{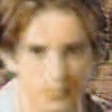} \hspace{-0.6em}
        \includegraphics[width=0.48\linewidth]{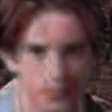} &  \vspace{-0.97\linewidth}
        \includegraphics[width=0.48\linewidth]{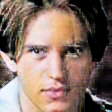} \hspace{-0.6em}
        \includegraphics[width=0.48\linewidth]{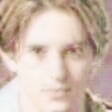} \hspace{-0.6em}
        \includegraphics[width=0.48\linewidth]{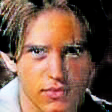} \hspace{-0.6em}
        \includegraphics[width=0.48\linewidth]{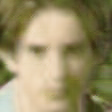} &  \vspace{-0.97\linewidth}
        \includegraphics[width=0.48\linewidth]{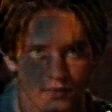} \hspace{-0.6em}
        \includegraphics[width=0.48\linewidth]{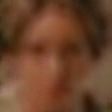} \hspace{-0.6em}
        \includegraphics[width=0.48\linewidth]{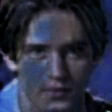} \hspace{-0.6em}
        \includegraphics[width=0.48\linewidth]{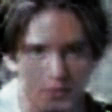} &  \vspace{-0.97\linewidth}
        \includegraphics[width=0.48\linewidth]{fig/generated/2_1.png} \hspace{-0.6em}
        \includegraphics[width=0.48\linewidth]{fig/generated/2_2.png} \hspace{-0.6em}
        \includegraphics[width=0.48\linewidth]{fig/generated/2_3.png} \hspace{-0.6em}
        \includegraphics[width=0.48\linewidth]{fig/generated/2_4.png} \\\midrule
        
        \includegraphics[width=0.97\linewidth]{fig/generated/3_0.png} & 
        \includegraphics[width=0.97\linewidth]{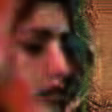} & \vspace{-0.97\linewidth}
        \includegraphics[width=0.48\linewidth]{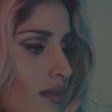} \hspace{-0.6em}
        \includegraphics[width=0.48\linewidth]{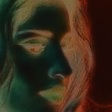} \hspace{-0.6em}
        \includegraphics[width=0.48\linewidth]{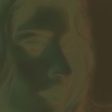} \hspace{-0.6em}
        \includegraphics[width=0.48\linewidth]{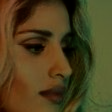} &  \vspace{-0.97\linewidth}
        \includegraphics[width=0.48\linewidth]{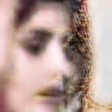} \hspace{-0.6em}
        \includegraphics[width=0.48\linewidth]{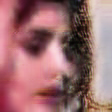} \hspace{-0.6em}
        \includegraphics[width=0.48\linewidth]{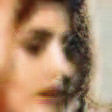} \hspace{-0.6em}
        \includegraphics[width=0.48\linewidth]{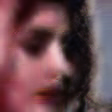} &  \vspace{-0.97\linewidth}
        \includegraphics[width=0.48\linewidth]{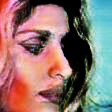} \hspace{-0.6em}
        \includegraphics[width=0.48\linewidth]{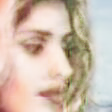} \hspace{-0.6em}
        \includegraphics[width=0.48\linewidth]{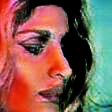} \hspace{-0.6em}
        \includegraphics[width=0.48\linewidth]{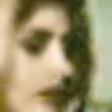} &  \vspace{-0.97\linewidth}
        \includegraphics[width=0.48\linewidth]{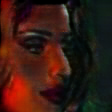} \hspace{-0.6em}
        \includegraphics[width=0.48\linewidth]{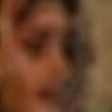} \hspace{-0.6em}
        \includegraphics[width=0.48\linewidth]{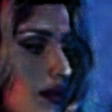} \hspace{-0.6em}
        \includegraphics[width=0.48\linewidth]{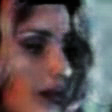} &  \vspace{-0.97\linewidth}
        \includegraphics[width=0.48\linewidth]{fig/generated/3_1.png} \hspace{-0.6em}
        \includegraphics[width=0.48\linewidth]{fig/generated/3_2.png} \hspace{-0.6em}
        \includegraphics[width=0.48\linewidth]{fig/generated/3_3.png} \hspace{-0.6em}
        \includegraphics[width=0.48\linewidth]{fig/generated/3_4.png} \\
    \bottomrule
    \end{tabularx}
    \caption{Ablation study of the augmentation network. Input images are shown in the first column. The subsequent columns show the results of different models trained without a certain module or loss. The texture style codes are randomly sampled from the normal distribution.}
    \label{fig:appendix_ablate_generated}
\end{figure*}

\begin{table*}[h]
\captionsetup{font=footnotesize}
\setlength{\tabcolsep}{5.0pt}
\begin{center}
\scriptsize
\begin{tabularx}{0.8\linewidth}{X || c|c|c|c|c || c|c|c|c|c|c|c|c}
\Xhline{2\arrayrulewidth}
\multirow{2}{*}{Model} & \multicolumn{5}{c||}{Modules}  & \multicolumn{3}{c|}{IJB-C (Vrf)} & \multicolumn{2}{c|}{IJB-C (Idt)} & \multicolumn{2}{c|}{IJB-S (V2S)} & LFW \\\cline{2-14}
& MM & $D_I$ & Rec & $D_Z$ & ND                                     & 1e-7 & 1e-6 & 1e-5 & Rank1 & Rank5& Rank1 & Rank5 & Accuracy \\\Xhline{2\arrayrulewidth}
 & & & & &                                                          & 72.74 & 85.33 & 90.52 & 94.99 & 96.75 & 56.35 & 66.77 & 99.82 \\\hline
(a) & & \checkmark & & & \checkmark                                   & 74.80 & 87.58 & 91.94 & 95.51 & 97.09 & 56.98 & 65.66 & 99.80\\\hline
(b) & \checkmark & & \checkmark & \checkmark & \checkmark             & 75.32 & 88.00 & 91.71 & 95.42 & 97.04 & 57.54 & 66.72 & 99.75 \\\hline
(c) & \checkmark & \checkmark & & & \checkmark                        & 74.51 & 87.49 & 91.97 & 95.61 & 97.18 & 57.17 & 66.24 & 99.78 \\\hline
(d) & \checkmark & \checkmark & \checkmark & & \checkmark             & 75.07 & 88.11 & 92.19 & 95.66 & 97.12 & 56.85 & 64.87 & 99.78 \\\hline
(e) & \checkmark & \checkmark & \checkmark & \checkmark &             & 73.99 & 86.52 & 91.33 & 95.33 & 97.04 & 58.47 & 66.00 & 99.73 \\\hline
(f) & \checkmark & \checkmark & \checkmark & \checkmark & \checkmark  & 77.39 & 87.92 & 91.86 & 95.61 & 97.13 & 57.33 & 65.37 & 99.75 \\\hline
\Xhline{2\arrayrulewidth}
\end{tabularx}\vspace{-0.5em}
\end{center}
\caption{Ablation study over different training methods of the augmentation network. ``MM'', ``$D_I$'', ``$D_Z$'', ``rec'', ``ND'' refer to ``Multi-mode'', ``Image Discriminator'', ``Reconstruction Loss'', ``Latent Style Discriminator'' and ``No Downsampling'', respectively. The first row is a baseline that uses only the domain adversarial loss but no augmentation network. ``Model (a)'' is a single-mode translation network that does not use latent style code.}
\label{tab:appendix_augmentation}
\end{table*}

\section{Ablation over the Settings of Augmentation Network}
In this section, we ablate over the training modules of the augmentation network. In particular, we consider to remove the following modules for different variants:
Latent-style code for multi-mode generation (MM), Image Discriminator ($D_I$), Reconstruction Loss (Rec), Style Discriminator ($D_z$ ) and the architecture without downsampling (ND). The qualitative results of different models are shown in Fig.~\ref{fig:appendix_ablate_generated}. Without the latent style code (Model a), the augmentation network can only output one deterministic image for each input, which mainly applies blurring to the input image. Without the image adversarial loss (Model b), the model cannot capture the realistic variations in the unlabeled dataset and the style code can only change the color channel in this case. Without the Reconstruction Loss (Model c), the model is trained only with adversarial loss but without the regularization of content preservation. And therefore, we see clear artifacts on the output images. However, adding reconstruction loss alone hardly helps, since the latent code used in the reconstruction of the unlabeled images could be very different from the prior distribution $p(z)$ that we use for generation. Therefore, similar artifacts can be observed if we do not add latent code adversarial loss (Model d). As for the architecture, if we choose to use an encoder-decoder style network as in the original MUNIT~\cite{MUNIT}, with downsampling and upsampling (Model e), we observe that the output images are always blurred due to the loss of spatial information. In contrast, with our architecture  (Model f), the network is capable of augmenting images with diverse color, blurring and illumination styles but without clear artifacts.

\begin{figure*}
    \centering
    \includegraphics[trim=0 340px 0 0, clip, width=\linewidth]{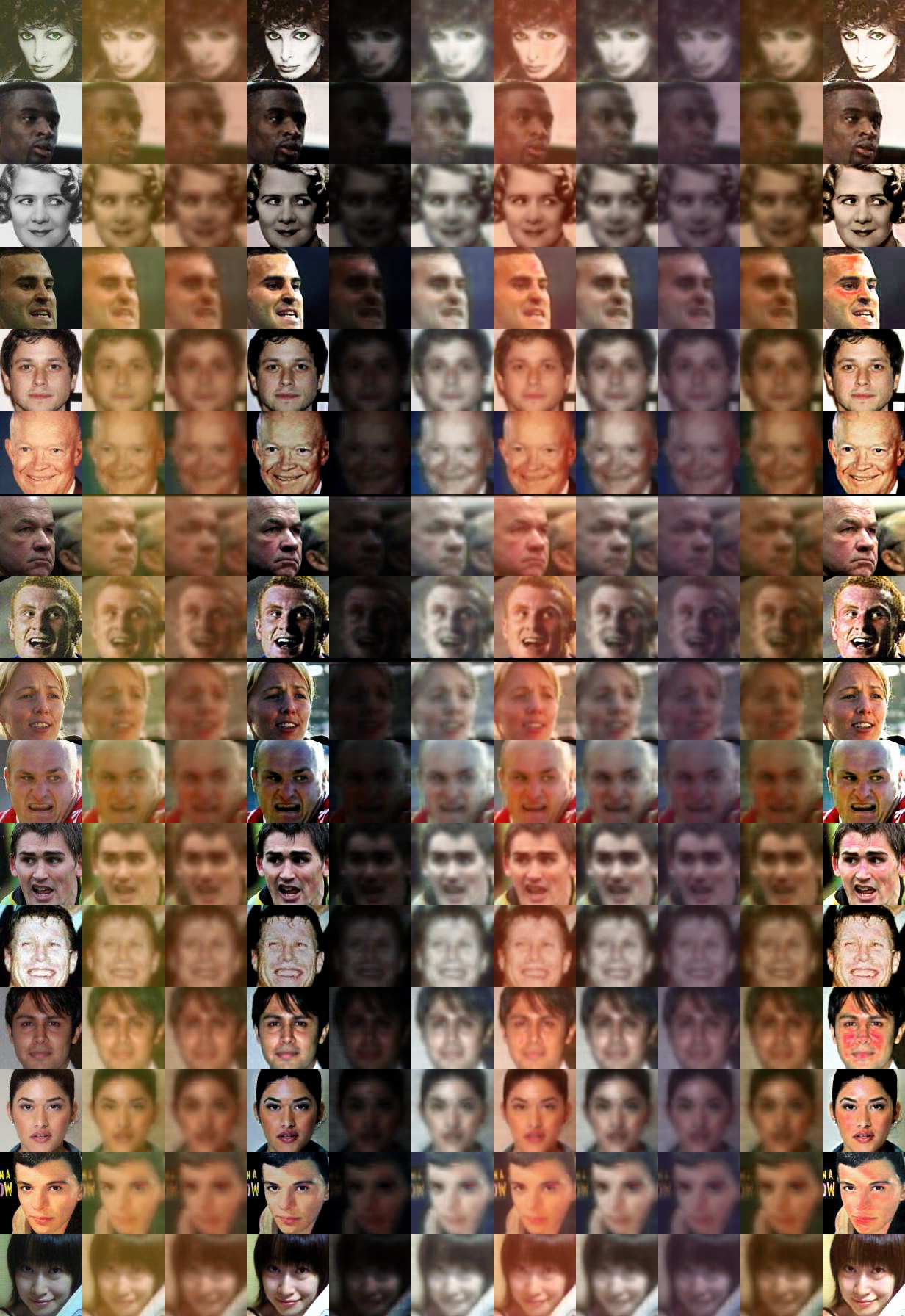}
    \caption{More examples of augmented images. The photos in the first column are the input images. The remaining images in each row are generated by the augmentation network with different style code.}
    \label{fig:appendix_generated}
\end{figure*}

Furthermore, we incorporate these different variants of augmentation networks into training and show the results in Table~\ref{tab:appendix_augmentation}. The baseline model here is a model that only uses domain alignment loss without augmentation network. In fact, compared with this baseline, using all different variants of the augmentation network achieves performance improvement in spite of the artifacts in the generated images. But a more stable improvement is observed for the proposed augmentation network across different evaluation protocols. We also show more examples of augmented images in Figure~\ref{fig:appendix_generated}.

\end{document}